\pdfoutput=1
\documentclass{article}

\usepackage{microtype}
\usepackage{graphicx}
\usepackage{subcaption}
\usepackage{booktabs} 
\usepackage{multirow}
\usepackage{tabularx}
\usepackage{color}
\usepackage{epsfig}
\usepackage{longtable}
\usepackage{lscape}
\usepackage{setspace}
\usepackage{appendix}
\usepackage{indentfirst}
\usepackage{hhline}
\usepackage{array}
\usepackage{caption}
\usepackage{subcaption}
\usepackage{floatrow}
\usepackage{booktabs}

\usepackage{hyperref}

\usepackage[accepted]{icml2025}

\usepackage{amsmath}
\usepackage{amssymb}
\usepackage{mathtools}
\usepackage{amsthm}

\usepackage[capitalize,noabbrev]{cleveref}


\theoremstyle{plain}

\theoremstyle{definition}

\theoremstyle{remark}

\usepackage[textsize=tiny]{todonotes}

\newcommand\blfootnote[1]{%
  \begingroup
  \renewcommand\thefootnote{}\footnote{#1}%
  \addtocounter{footnote}{-1}%
  \endgroup
}

\icmltitlerunning{Online Curvature-Aware Replay}

\begin{document}

\twocolumn[
\icmltitle{Online Curvature-Aware Replay: \\ Leveraging $\mathbf{2^{nd}}$ Order Information for Online Continual Learning}



\icmlsetsymbol{equal}{*}

\begin{icmlauthorlist}
\icmlauthor{Edoardo Urettini}{equal,unipi, sns}
\icmlauthor{Antonio Carta}{equal,unipi}
\end{icmlauthorlist}




\vskip 0.3in
]




\begin{abstract}
Online Continual Learning (OCL) models continuously adapt to nonstationary data streams, usually without task information. These settings are complex and many traditional CL methods fail, while online methods (mainly replay-based) suffer from instabilities after the task shift. To address this issue, we formalize replay-based OCL as a second-order online joint optimization with explicit KL-divergence constraints on replay data. We propose Online Curvature-Aware Replay (OCAR) to solve the problem: a method that leverages second-order information of the loss using a K-FAC approximation of the Fisher Information Matrix (FIM) to precondition the gradient. The FIM acts as a stabilizer to prevent forgetting while also accelerating the optimization in non-interfering directions. We show how to adapt the estimation of the FIM to a continual setting stabilizing second-order optimization for non-iid data, uncovering the role of the Tikhonov regularization in the stability-plasticity tradeoff.
  Empirical results show that OCAR outperforms state-of-the-art methods in continual metrics achieving higher average accuracy throughout the training process in three different benchmarks.
\end{abstract}

\blfootnote{$^*$ Equal contribution; $^1$ University of Pisa, Pisa, Italy; $^2$ Scuola Normale Superiore, Pisa, Italy}
\blfootnote{Contacts: edoardo.urettini@sns.it; antonio.carta@unipi.it}

\section{Introduction}
Online Continual Learning (OCL) models are trained continuously on a nonstationary data stream. The goal is to obtain a model that is accurate at any point in time, quickly adapts to new data (i.e. plasticity), and does not forget old information (i.e. stability), without any information about task start, end, or identity. In this setting, many standard Continual Learning (CL) approaches, cannot be applied \cite{aljundi2019task,mai2022online}. 
Many specific OCL methods have been developed, nearly all replay-based \cite{yoo2024layerwise}, but several shortcomings still exist: \cite{DBLP:conf/iccvw/Soutif-Cormerais23} shows that most OCL methods have high forgetting and fail to beat a simple replay baseline on some metrics; \cite{lirias4071238,DBLP:conf/iclr/CacciaAATPB22} discovered the stability gap, a sudden drop in performance at task boundaries; \cite{DBLP:journals/nature/DohareHLRMS24} even observed loss of plasticity, limiting the ability to learn in time.

These results suggest fundamental failures of the algorithms. Most methods focus on preventing forgetting at the end of a learning task, an approach that does not ensure stable optimization throughout the entire stream (stability gap); furthermore, stability is often optimized at the expense of plasticity. We argue that a proper CL optimizer should seek to maximize both stability and plasticity, assuming that the model is large enough, at every step in time, interpreting it as a continual filtering process (filter intended as in nonstationary time series literature \cite{durbin2012time}).

This paper proposes Online Curvature-Aware Replay (OCAR), a novel method designed for replay-based online continual learning, aiming at tackling the challenges in both plasticity and stability inherent in this scenario with a continual optimization approach at every step in time.
We formalize OCL as the joint optimization on past and new data, with past data approximated using a limited replay buffer, and adding an explicit constraint on the variation of KL-divergence on previous information. The Fisher Information Matrix (FIM) is used to capture the loss function's curvature, providing both plasticity and stability constraints in the model distribution space.
When the KL divergence is used as a metric, the FIM has the additional value of describing the curvature of the parameter space itself, being the Riemannian metric tensor of that space \cite{amari2016information}. We can directly adapt our gradient to the geometry of the space in stark difference with traditional CL methods that uses the FIM as a penalization term \cite{kirkpatrick2017overcoming}. 
Kronecker-factored Approximate Curvature (K-FAC) \cite{martens2015optimizing} is used to efficiently approximate the FIM, with some critical adjustments to make it work in OCL settings.

The main contributions of this paper are: the design of Online Curvature-Aware Replay (OCAR) as a combination of replay, second-order optimization, and information geometry; the analysis of the Tikhonov regularization and its ratio with the learning rate in the stability-plasticity tradeoff; an improvement of state-of-the-art performance in standard computer vision benchmarks across all the stability and plasticity metrics.

\section{Related Work}
\textbf{Continual Stability and OCL:} Most continual learning methods assume that stability must be kept at the expense of plasticity~\cite{DBLP:journals/pami/LangeAMPJLST22,DBLP:journals/pami/MasanaLTMBW23}, the so called plasticity-stability tradeoff, and therefore are designed to preserve knowledge about previous tasks to mitigate catastrophic forgetting~\cite{french1999catastrophic}. However, recent evidence in \cite{DBLP:conf/iclr/LangeVT23,DBLP:conf/iclr/CacciaAATPB22} suggest that even the methods with high "stability" measured at the end of tasks suffer from high instability and forgetting immediately after the task switch, and they recover the lost performance over time. \cite{DBLP:journals/corr/abs-2406-05114} found evidence of the stability gap even in incremental i.i.d. settings. In OCL settings, \cite{DBLP:conf/iccvw/Soutif-Cormerais23} showed that some state-of-the-art methods are unable to outperform a simple reservoir sampling baselines on some fundamental stability metrics. 
Furthermore, CL methods also fail at keeping plasticity, and \cite{DBLP:journals/nature/DohareHLRMS24} provides evidence of the loss of plasticity in deep continual networks. Overall, the literature suggests that CL methods fail at both stability and plasticity due to instabilities in the learning dynamics. Recently, some methods such as OnPro~\cite{wei2023online} and OCM~\cite{guo2022online} proposed novel self-supervised auxiliary losses and prototype-based classifiers, two approaches orthogonal to our optimization-based method.

\textbf{Optimization in Continual Learning:} Most CL optimization algorithms are designed to prevent forgetting by removing interfering updates. GEM \cite{DBLP:conf/nips/Lopez-PazR17,DBLP:conf/iclr/ChaudhryRRE19} models interference using the dot product of the task gradients and constrains the model updates to have positive dot products with the gradients of previous tasks. Subsequent work explored orthogonal projection methods~\cite{DBLP:conf/iclr/SahaG021,DBLP:conf/aistats/FarajtabarAML20} that either extend the idea of interfering gradients or project in the null space of the latent activations. \cite{DBLP:conf/nips/MirzadehFPG20} discusses the relationship between the curvature of the first task and the forgetting, proposing a hyperparameter schedule that implicitly regularizes the curvature. More recently, LPR\cite{yoo2024layerwise} exploits proximal optimization in the L2 space of latent activations, and it is the only projection-based optimizer compatible with replay. \cite{DBLP:journals/corr/abs-2311-04898} proposes a combination of GEM and replay as a potential mitigation for the stability gap.

\textbf{Natural Gradient and FIM in CL:} Natural gradients can be used to train neural networks thanks to efficient approximations of the Fisher Information Matrix (FIM)~\cite{martens2012training}, such as the K-FAC~\cite{martens2020new} and E-KFAC~\cite{DBLP:conf/nips/GeorgeLBBV18}. Interestingly, \citet{DBLP:conf/icml/Benzing22} showed that K-FAC seems to work better than the full FIM in some empirical settings, which is connected to a form of gradient descent on the neurons. In continual learning, the FIM is typically used to approximate the posterior of the weights with a Laplace approximation. The result is a quadratic regularizer that is combined with the loss on new data, as introduced by EWC~\cite{kirkpatrick2017overcoming} and its several extensions~\cite{DBLP:conf/eccv/ChaudhryDAT18,liu2018rotate,DBLP:journals/corr/abs-1712-03847}. \citet{DBLP:conf/iclr/MagistriTS0B24} proposed to use the FIM only for the final layer, which can be computed efficiently with a closed-form equation. As an alternative, FROMP~\cite{pan2020continual} computes a Gaussian Process posterior, which is also used to estimate the importance of samples in the replay buffer. \citet{daxberger2023improving} proposes a method that combines EWC, replay, and knowledge distillation. More relevant to our work, NCL~\cite{DBLP:conf/nips/KaoJVBH21} proposed a modified natural gradient step with a quadratic posterior as in EWC. Here the Fisher is computed only at the boundaries and the method does not support rehearsal, making it difficult to implement in OCL.

\textbf{Comparison with our work} Most CL methods use the FIM to compute a Laplace approximation of the posterior. This is not possible in OCL because the model is never assumed to be at a local minimum, not knowing task boundaries or length. In general, in the CL literature, the FIM is often restricted to its use in quadratic penalties. While quadratic regularizers are easier to use, we show that the use of the Fisher as a gradient preconditioner is a promising direction, improving the optimization path. Many of the limitations found for EWC and similar methods may be due to some suboptimal choices in the use of the FIM, such as the use of the empirical FIM, popular in the CL literature but with different properties from the FIM~\cite{kunstner2019limitations}. Another difference with the literature is that our optimizer is compatible with replay, unlike most projection-based methods. Furthermore, while most methods penalize plasticity indirectly to prevent forgetting, our approach is designed to improve both on learning speed and forgetting.

\section{Online Continual Learning}
In continual learning (CL), the model learns incrementally from nonstationary data. 
In most CL settings, we can identify a sequence of tasks $\mathcal{T}_1, \hdots, \mathcal{T}_N$, each one with its own distribution. For example, in class-incremental learning \cite{DBLP:journals/corr/abs-1904-07734}, each tasks has different classes, and tasks are seen sequentially during training. The goal of the model is to learn all the tasks seen during training. Given the model parameters $w_t$ learned at time $t$, a loss function $\mathcal{L}$, and a test set for each task $\mathcal{D}_1, \hdots, \mathcal{D}_N$, we can evaluate the model with the average task loss $\mathcal{L}^{avg}(t) = \sum_{i=1}^N \mathcal{L}(w_t, \mathcal{D}_i)$.

Online continual learning (OCL) requires some additional constraints and desiderata~\cite{DBLP:conf/iccvw/Soutif-Cormerais23,yoo2024layerwise,DBLP:journals/ijon/MaiLJQKS22}: (\emph{D1-Online training}) at each step, we do not have access to the whole training dataset for the current task, but only a small minibatch, which can be processed for a limited amount of time; (\emph{D2-Anytime inference}) the model should be ready for inference at any point in time; Consequently, (\emph{D3-Continual stability}) the model must be stable at any point in time, instead of only at the task boundaries; (\emph{D4-fast adaptation}) the model must also be able to learn quickly from new data.


Following the literature, we focus on replay-based methods, which use a limited buffer of previous data for rehearsal, by combining new minibatches with samples from the buffer at each iteration. This work assumes no access to knowledge about task identities, boundaries, or length. Each single observation in the stream can be sampled from different distributions/tasks.


\section{Online Curvature-Aware Replay}
We show the building process of our method, starting from the current ER optimization and expanding it to a second-order method, then approximated with FIM and making some final adjustments to improve its CL performance.
\subsection{The optimization problem}
Online continual learning (OCL) is an online learning problem in a nonstationary setting. Common optimizers in machine learning implicitly assume stationarity, which justifies the gradient estimate from the minibatches (e.g.: ADAM \cite{DBLP:journals/corr/KingmaB14}, SGD). Instead, in OCL the distribution can change at any point in time (non i.i.d.). 
At each step, the method can only use the current minibatch and, in replay-based methods, an additional minibatch sampled from a small buffer of old data. 

\textbf{First-order optimization:}
We define our learning process as a sequence of local optimization problems solved at each step. Differently from stationary settings, these problems can be weakly dependent one on another, preventing the use of more "global" approaches (e.g, learning rate decay and momentum~\cite{lecun2015deep}). Each single step must be meaningful by itself. What is forgotten or not learned could be lost forever.

The Kullback–Leibler (KL) divergence \cite{thomas2006elements,lecun2015deep} is used as our objective, aiming to minimize the "distance" between the predicted and the real distribution. The KL is estimated on the current batch of data $\hat{KL}(y_D || f_w(x_D)) = \frac{1}{N} \sum_{i}^N KL(\hat{p}(y_{D, i}| f_{w^*}(x_{D, i})|| p(y_{D, i}| f_{w}(x_{D, i}))$. In our notation $D$ represents the batch of data, $i$ the specific sample, $y$ the target variable dependent on the observed $x$. $f_w$ is the model parametrized by $w$ that is supposed to represent the reality when $w=w^*$.The empirically observed distribution is $\hat{p}(y_{D, i}| f_{w^*}(x_{D, i})$.

Experience Replay (ER) \cite{chaudhry2019continual} keeps a small buffer $\mathcal{B}$ of past data and solves a joint optimization on a batch $N_t$ sampled from the unknown real current data distribution and a batch $B_t$ sampled from $\mathcal{B}$ :
\begin{equation}
\label{eq: ER_problem}
\begin{aligned}
& \min_{\delta_t} \quad \hat{KL}(y_{N_t} \, || \, f_{w_t}(x_{N_t})) +  \hat{KL}(y_{B_t} \, || \, f_{w_t}(x_{B_t}))\\
& \text{subject to} \quad \frac{1}{2}||\delta||_2^2 \leq \epsilon, 
\end{aligned}
\end{equation}
where $\delta_t = w_t - w_{t-1}$. Here and throughout the paper we write a sum of the two KL divergences for visual ease. Usually (and we also do this in practice) the mean is taken, combining the data in a single batch with no distinctions. The KL divergence is approximated by Taylor expansion:
\begin{align*}
    \hat{KL}(y_{D} \, || \, f_w(x_{D})) \approx  \; &\hat{KL}(y_{D} \, || \, f_{w = w_0}(x_{D})) + \\ &\nabla_{t}^T \delta_t + \delta_t^T \mathbf{H}_{t} \delta_t
\end{align*}
where $\nabla_t = \nabla_w \hat{KL}(y_{N_t} \, || \, f_{w = w_0}(x_{N_t}))$ is the gradient of the model on the current data, and $\mathbf{H}_{t} = \mathbf{H}_w \hat{KL}(y_{N_t} \, || \, f_{w = w_0}(x_{N_t}))$ the Hessian. $D$ can be both $N_t$ or $B_t$. After approximation, the solution for problem \ref{eq: ER_problem} is $\delta_t^* = -\frac{1}{\lambda} (\nabla_{N_t} + \nabla_{B_t})$ where $\lambda$ is the Lagrange multiplier of the constraint. ER actively optimizes on the buffer and the current data distributions with no distinctions and no forgetting constraint. The only stability requirement is $\frac{1}{\lambda}$ (the learning rate), limiting the movement of the weights in all directions, trading stability for plasticity.

\textbf{Failure of ER with first-order optimization:}
Unfortunately, the first-order information can be a poor approximation in CL. Often, information about the curvature is necessary to avoid catastrophic forgetting. For example, at task boundaries, the model is often close to a minimum for the previous task. In that case, buffer gradients $\nabla_B \approx 0$, while new gradients $\nabla_N$ can potentially be much higher, dominating the update direction. This issue partially explains the stability gap \cite{lirias4071238}.
Second-order methods can solve this problem by enlarging the "sight" of our optimizer with the local variation of the variation.
Using a second-order Taylor expansion problem \ref{eq: ER_problem} is solved by
\begin{equation*}
    \delta_t^* = -(\mathbf{H}_{N_t} + \mathbf{H}_{B_t} + \lambda \mathbf{I})^{-1}(\nabla_{N_t} + \nabla_{B_t}),
\end{equation*}
where the Lagrange multiplier $\lambda$, still related to the $L2$ regularization imposed on the update, now acts as a Tikhonov damping term \cite{martens2012training}. Unlike the learning rate in SGD, $\tau$ has a different effect on the eigendirections of the Hessian, which depends on their eigenvalues (see Sec. \ref{sec:car_stability}). 

\textbf{Stability constraint:}s approach already improves plasticity (D4), accelerating learning on both current and past data through Newton optimization, it still lacks an explicit stability constraint (D3) necessary for non-i.i.d. settings. Combining experience replay, second-order methods, an explicit requirement for stability on past data and $L2$ regularization on the update the problem becomes:
\begin{equation*}
\begin{aligned}
\min_{\delta} \quad &\hat{KL}(y_{N_t} \, || \, f_{w_t}(x_{N_t})) +  \hat{KL}(y_{B_t} \, || \, f_{w_t}(x_{B_t}))\\
\text{subject to}\quad  &\hat{KL}(f_{w_{t-1}}(x_{B_t}) \, || \, f_{w_t}(x_{B_t})) \leq \rho\\
& \frac{1}{2}||\delta||_2^2 \leq \epsilon. 
\end{aligned}
\end{equation*}
The term $\hat{KL}(f_{w_{t-1}}(x_{B_t}) \, || \, f_{w_t}(x_{B_t}))$ measures the variation of the model predicted distribution on buffer data, and its expansion around the pre-update parameters $w_{t-1}$ will have zero zeroth and first-order terms. The remaining term is controlled by the Hessian of the KL-divergence evaluated at $w=w_{t-1}$, which is exactly the Fisher Information Matrix \cite{ollivier2017information} \cite{martens2020new}:
\begin{equation*}
    \mathbf{F}_{B_t, ij} = \left. \frac{\partial^2 \hat{KL}(f_{w_{t-1}}(x_{B_t}) \, || \, f_{w}(x_{B_t}))}{\partial w_i \, \partial w_j} \right|_{w_t = w_{t-1}}.
\end{equation*}
The optimization problem becomes
\begin{equation}\label{eq:complete_opt}
\begin{aligned}
\min_{\delta} \quad & \nabla_{N_t}^T \delta +  \delta^T \mathbf{H}_{N_t} \delta +  \nabla_{B_t}^T \delta +  \delta^T \mathbf{H}_{B_t} \delta\\
\text{subject to}\quad &\frac{1}{2}\delta^T \mathbf{F}_{B_t} \delta \leq \rho &\ \\
&\frac{1}{2}||\delta||_2^2 \leq \epsilon,
\end{aligned}
\end{equation}
which is solved by
\begin{equation*}
    \delta_t^* = -(\mathbf{H}_{N_t} + \mathbf{H}_{B_t} + \lambda \mathbf{F_{B_t}}+\tau \mathbf{I})^{-1}(\nabla_{N_t} + \nabla_{B_t}),
\end{equation*}
where $\lambda$ controls the strength of the stability constraint, ensuring that the predictions of the model remain stable, while $\tau$ is the Tikhonov regularization controlling directions of maximum acceleration. As done in many second-order methods~\cite{martens2020new,martens2012training}, a learning rate $\alpha$ can be used to uniformly control the step size in all directions, making the final update be $\alpha \delta_t^*$. As we will show, the relation between $\alpha$ and $\tau$ has an important impact on the learning dynamics and the stability-plasticity tradeoff. 

\subsection{Estimations and approximations}
\textbf{Approximating the Hessians: }Computing two Hessians and one Fisher Information Matrix would be impractical and inverting them unfeasible. Luckily, in the second-order optimization literature, it has been shown how for some distributions (including the multivariate normal and the multinomial), the FIM is equivalent to a Generalized Gauss-Newton (GGN) matrix, both approximating the Hessian of the KL-divergence \cite{martens2020new}. They are equal to the Hessian when the model describes well the data (near optimum) and the work of Martens \cite{martens2020new} illustrates some general advantages of these approximations over the Hessian also when far from optimum.

Moreover, the use of the FIM has the additional interpretation of describing the curvature of the parameter space itself. In this space, each point represents a distribution, requiring the use of the KL divergence as "distance". The KL-divergence of infinitesimal displacements corresponds to $\frac{1}{2} F_{ij} d w_i d w_j$  giving rise to the Fisher Information as the metric tensor used in this manifold \cite{amari2016information}. More than a simple approximation of the Hessian, the FIM can be used to correctly measure distances in the manifold where we are performing the gradient descent if the steps are small enough. Assuming a small learning rate, our method would use the preconditioner not as in a Newton method (that usually requires a learning rate of 1 to directly go the the approximate optimum) but as a metric adaptation. 

We believe these two results justify the use of the Fisher information in nonstationary settings as OCL. Following this, we approximate the two Hessians matrices of our solution with the FIM, greatly simplifying the computations. The new optimal update becomes (including the learning rate):
\begin{equation*}
\label{eq:OCAR}
    \delta_t^* = -\alpha (\mathbf{F}_{N_t} + (1 + \lambda) \mathbf{F_{B_t}}+\tau \mathbf{I})^{-1}(\nabla_{N_t} + \nabla_{B_t}).
\end{equation*}
This implies the second-order information is now taken with respect to the model prediction instead of the real targets as also $\mathbf{F}_{N_T} = \mathbf{H}_w \; \hat{KL}(f_{w_{t-1}}(x_{N_t}) || f_{w}(x_{B_t})$. The gradient of our method corresponds to the usual gradient obtained from the loss function when the loss is a derivation of the KL divergence (negative log-likelihood, cross-entropy, etc...). The weighted sum (or mean) of the two FIMs can now be obtained by computing a single FIM on the batch, giving more weight to the buffer data.

There is a deep connection between our method and Natural Gradient (NG) \cite{amari1998natural}. We have to underline that our building process is much different from the one used for the original NG, thought for stationary settings. Moreover, our preconditioner is not really the FIM of the model, but a modified and regularized version stemming from our OCL second-order optimization with the use of replay. We then cannot assume all benefits of NG would apply to our case. 

\textbf{The Empirical FIM: } The Fisher is also defined as the expected value of the squared score $F = \sum_n \mathbb{E}_{y \sim p(y|f_w(x_n))}[\nabla_w \log p(y|f_w(x_n)) \nabla_w \log p(y|f_w(x_n))^T ]$. One could argue that a better preconditioner for the gradient would be the Empirical Fisher (EF) matrix, computed using the real target $y$ instead of the one sampled from the predicted distribution, in particular for approximating the Hessian of new data $H_{N_t}$. Besides getting us outside the theory about Natural Gradient and Fisher/GGN equivalence, it has been shown it is a questionable choice, even when the model is not a good description of the data \cite{kunstner2019limitations}. For these reasons, OCAR uses the "real" Fisher, unlike other traditional CL approaches \cite{kirkpatrick2017overcoming}. 

\textbf{K-FAC: }This approach has theoretical advantages, but it requires the inversion of an extremely large matrix, unfeasible for large networks. While diagonal approximations are frequently used~\cite{kirkpatrick2017overcoming,kingma2014adam}, given the particular challenges of OCL a more informative approximation is needed. We rely on the Kronecker-factored Approximate Curvature (K-FAC) in its block-diagonal version to approximate the FIM \cite{martens2015optimizing}:
\begin{equation*}
    \Tilde{F} = \text{diag}(\bar{A}_{0,0} \otimes G_{1,1}, ... ,\bar{A}_{l-1,l-1} \otimes G_{l,l}),
\end{equation*}
where each block corresponds to an approximation of the covariance matrix of the score of a specific layer, obtained by the Kronecker product of $\bar{A}_{i, i} = \mathbb{E}[\bar{a}_i \bar{a}_i^T]$ and $G_{i, i} = \mathbb{E}[g_i g_i^T]$. $\bar{a}_i$ is the vector of the layer activations (with an additional $1$ appended for the bias) and $g_i$ the gradient of the prediction with respect to the output of the layer before the activation function. The expected values of both are computed with an Exponential Moving Average (EMA) of past values. The efficiency of this method relies on the property of the Kronecker product $(A \otimes B)^{-1} = A^{-1} \otimes B^{-1}$ allowing us to invert much smaller matrices for each layer, ignoring layer interactions. 

\subsection{Nuts and Bolts for OCL}\label{sec:cl_nuts}

To make the method work in practice, we found some adjustments are needed. 

\textbf{Hyperparameter optimization and $\tau$ scheduling:} Usually, hyperparameter selection is done only on the first $K$ tasks of the stream, but it must generalize to longer streams during training. Our method uses three hyperparameters: the learning rate $\alpha$, the Tikhonov regularizer $\tau$, and the parameter $\gamma$ for the EMA used for Kronecker factors estimation. In hyperparameter selection, we found it beneficial to search for a value for the increase of $\tau$ instead of $\tau$ itself. $\tau$ is then initialized at the same value of the learning rate and increased by the selected value at each optimization step improving long-term stability. 

\textbf{Estimate of K-FAC factors at boundaries:} In class-incremental settings, the shape of the classifier will grow over time as new classes are observed. This means the K-FAC factor $G_{l,l}$ of the last layer will change shape, also breaking the relations with the previously observed gradients. To avoid keeping track of errors, if $G_{l,l}$ changes shape, the EMA of this factor is reset. This is not done for the other factor of the last layer $\bar{A}_{l-1,l-1}$. We assume the model has consistent representations, and that the majority of instability is happening on the classifier. 

\textbf{Scheduling of $\lambda$:} The stability constraint of problem \ref{eq:complete_opt} is estimated on the current batch extracted from the buffer, but it should represent the whole buffer. As new experiences are encountered in time, the information content of the buffer will grow, with less redundancy. At the limit, we can get a buffer where each example represent a different class or domain. Being $\lambda$ the Lagrange multiplier, it is directly connected with the constraint $\rho$. By decreasing $\rho$ to have a stronger constraint and save buffer information, we get an increase in $\lambda$. For class-incremental problems, $\lambda$ increases with the number of different classes encountered. In domain-incremental settings, it grows in time (as done with $\tau$). 

The composition of all these pieces together forms the \textbf{Online Curvature-Aware Replay (OCAR)}: at each step, the method is the optimal step of a second-order replay-based optimization problem, approximated with the FIM made computationally viable via K-FAC, with some fundamental adjustments to make it work in practice. The algorithm can be found in the Appendix \ref{app: algo}.

\begin{figure}[t]
    \centering
    \begin{subfigure}[t]{0.95\textwidth}
        \centering
        \includegraphics[width=\textwidth]{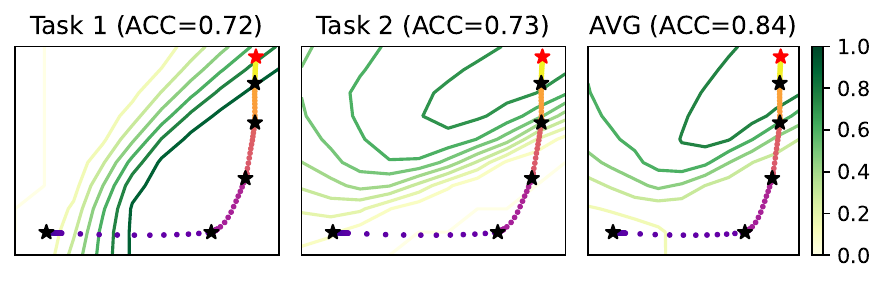}
        \caption{OCAR 2D projection of the learning trajectory.}\label{fig:proj_car}       
    \end{subfigure}
    \begin{subfigure}[t]{0.95\textwidth}
        \centering
        \includegraphics[width=\textwidth]{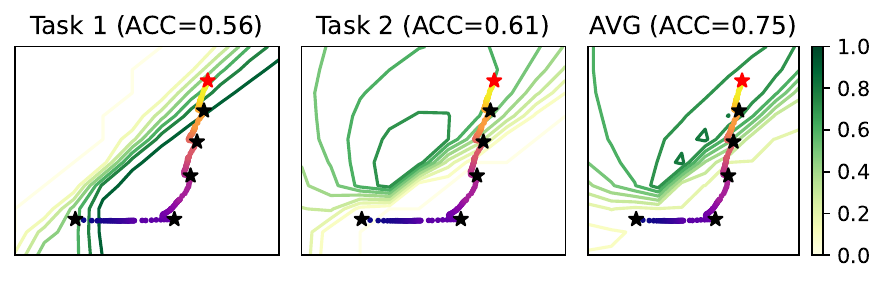}
        \caption{ER 2D projection of the learning trajectory.}\label{fig:proj_er}       
    \end{subfigure}
    \caption{2D projections of the training trajectories for ER and OCAR on Split MNIST (5 Tasks). Loss surface on the first task (left), second task (middle), and the average loss on all the $5$ tasks (right). The black stars highlight the task boundaries. More details on the 2D projections and additional plots are available in the Appendix.}\label{fig:traj}
\end{figure}

\section{Continual Stability in OCAR}\label{sec:car_stability}

In this section, we provide a qualitative analysis in a simplified setting, showing how OCAR results in a smoother continual optimization compared to ER and how $\alpha$, $\tau$, and their ratio can be used to control stability and plasticity.

\textbf{Loss landscape and model trajectories:} We can visualize the improvements in the optimization trajectory of OCAR in a simple continual learning setting. We train a small feed-forward network with ER and OCAR on Split MNIST (5 Tasks). Given the small size of the model, we can store the entire training history of the model, which allows us to plot 2D projections of the model trajectory in the loss landscape (Figure \ref{fig:traj}) (details in Appendix \ref{app: trajectory}).

Looking at their task-wise and joint loss surfaces in figures \ref{fig:proj_car} and \ref{fig:proj_er}, OCAR shows a much smoother model trajectory across all analyzed loss landscapes (Task 1, Task 2, and Average loss), which results in consistent improvements over time (plasticity) and mild forgetting (stability). Second-order information moves the optimization directly toward the next minimum (black stars in the plots), in fewer steps. On the other hand, ER always suffers from instability at the task boundaries (right after black stars in the plots) which results in an abrupt deviation from the optimal path. The learning curves on the first task (Fig. \ref{fig:curve_t0}) and the average of all tasks (Fig. \ref{fig:curve_joint}), available in Appendix \ref{app: figures}, confirm the result. OCAR maintains a smoother learning path without experiencing any stability gap on the first task and ending the stream with higher overall accuracy. ER instead, while still performing well in the basic MNIST setting, suffers from much more instability during training.

\textbf{Role of the Eigenvectors and Hyperparameters:} One approach to understand OCAR effect is to study how the eigenvalues of the matrix $\alpha (F_N + (1 + \lambda) F_B + \tau I)^{-1}$ are related to the hyperparameters $\alpha$, $\lambda$, and $\tau$. We can diagonalize  $\bar{F} = F_N + (1 + \lambda) F_B = Q \Sigma Q^T$, where $Q$ is a unitary matrix where the rows are the eigenvectors and $\Sigma$ a diagonal matrix with the eigenvalues $\sigma_i$. Therefore, we find that the eigenvalues of $\alpha (F_N + (1 + \lambda) F_B + \tau I)^{-1}$ are 
$\sigma_i^* = \frac{\alpha}{\sigma_i + \tau}.$
In the new coordinate system defined by the eigenvectors $Q$, we can interpret OCAR as slowing or accelerating directions depending on their curvature. Directions with $\sigma^*_i \ll 1$ ($\sigma^*_i \gg 1$) correspond to directions with high (low) curvature for some tasks.
The learning rate $\alpha$ and the Tikhonov regularization $\tau$ rescale these eigenvalues (as shown in Figure \ref{fig:lr_ratio_curve} in Appendix).  In particular, a learning rate $\alpha < 1$ decreases the step size. Conversely, $\tau$ mitigates the acceleration caused by small eigenvalues. When $\sigma_i \rightarrow 0$,  $\sigma^*_i \rightarrow \frac{\alpha}{\tau}$, limiting the maximum acceleration in each direction. When $\sigma_i \gg \tau$, we have $\sigma^*_i \approx \frac{\alpha}{\sigma_i}$, which is approximately independent of $\tau$.

Empirically, we find that the spectrum spans several orders of magnitude, with the smallest eigenvalues close to zero and a small set of very large values around $(10^4, 10^6)$ (consistently with the common intuition behind methods such as EWC, which expect few important parameters).  

\begin{figure}[t]
    \begin{subfigure}[b]{0.45\textwidth}
        \centering
        \includegraphics[width=\textwidth]{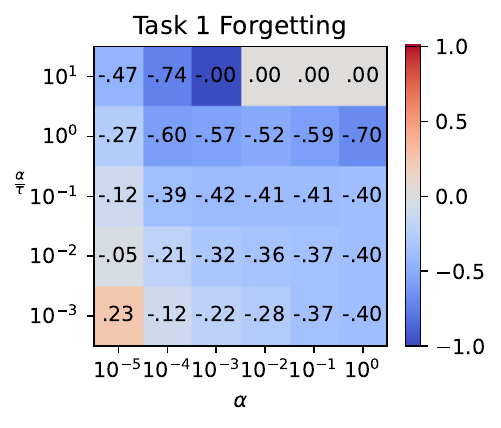}
    \end{subfigure}
    \begin{subfigure}[b]{0.45\textwidth}
        \centering
        \includegraphics[width=\textwidth]{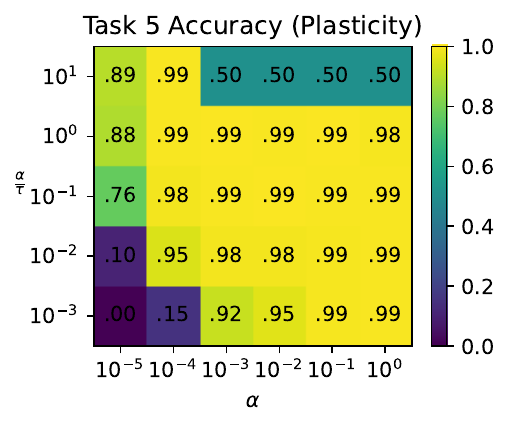}
    \end{subfigure}

    \caption{Grid search over $\alpha$ and $\frac{\alpha}{\tau}$: (left) forgetting on the first task, (right) plasticity measured as the accuracy on the final task. Metrics are computed on the test stream at the end of training.}\label{fig:lr_ratio_grid}
\end{figure}

\textbf{Stability-Plasticity tradeoff of OCAR hyperparameters:}
An interpretation of OCAR hyperparameters in the stability-plasticity tradeoff can be given.
While higher $\lambda$ gives more importance to the FIM of the buffer (stability) and higher values of $\alpha$ allow larger learning steps (plasticity), the role of $\tau$ and $\frac{\alpha}{\tau}$ is less intuitive. First, higher values of $\tau$ are needed for the introduction of new classes, that creates instability due to the FIM being the variance of gradients (See the Appendix \ref{app: Fisher computations} for a detailed explanation). Figure \ref{fig:lr_ratio_grid} shows the results of a grid search on $\alpha$ and $\frac{\alpha}{\tau}$ (Complete figure in Appendix fig \ref{fig:lr_ratio_grid_complete}). We can make some empirically-based speculations: (1) the ratio $\frac{\alpha}{\tau}$ controls the learning plasticity. As a result, bottom-left elements in Figure \ref{fig:lr_ratio_grid} show less plasticity; (2) $\frac{\alpha}{\tau}$ controls the effective step size in the eigendirections with small eigenvalues, problematic at task boundaries. If the ratio is too low, OCAR may use updates that are too large, resulting in instability (top-right); (3) Decreasing $\frac{\alpha}{\tau}$ consistently increases forgetting; (4) After a sufficient number of steps after a task drift, the FIM becomes more stable, and $\frac{\alpha}{\tau}$ becomes less important in the learning dynamics. At this point, small $\alpha$ will result in conservative updates, with low plasticity and high stability. In the figure, we notice that for equal values of the ratio, models may have a similar accuracy but a different stability-plasticity tradeoff, depending on their $\alpha$ (low=stability, high=plasticity). 


\begin{figure}[t]
    \centering
    \includegraphics[width=1\linewidth]{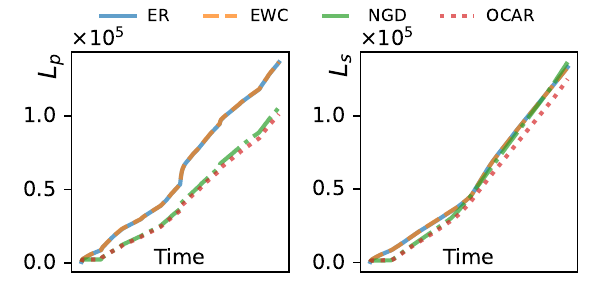}
    \caption{\textit{Left}: $L_p$ Cumulative loss of single batches. \textit{Right}: $L_s$ Cumulative loss measured on all previous data of the stream.}
    \label{fig:qualitative}
\end{figure}

\begin{table*}[t]
    \caption{Results on Split CIFAR100 (20 Tasks) and Split Tiny ImageNet (20 Tasks). Best in bold for the base methods. Best underlined for all methods included OCAR+OTHER.}\label{tab:tinyimnet}
    \resizebox{1.0\textwidth}{!}{%
    \begin{tabular}{l|cccc|cccc}
    \hline \multirow[t]{2}{*}{Method} & \multicolumn{4}{c}{Split-Cifar100 (20 Tasks)} & \multicolumn{4}{c}{Split-TinyImagenet (20 Tasks)} \\
    & Acc $\uparrow$ & $A A A^{\text {val }} \uparrow$ & WC-Acc ${ }^{\text {val }} \uparrow$ & Probed Acc $\uparrow$ & Acc $\uparrow$ & $A A A^{\text {val }} \uparrow$ & WC-Acc ${ }^{\text {val }} \uparrow$ & Probed Acc $\uparrow$ \\
    \midrule
    i.i.d & $35.3 \pm 1.5$ & - & - & $45.8 \pm 0.6$ & $26.5 \pm 0.6$ & - & - & $34.3 \pm 0.5$ \\ \midrule
    ER\cite{chaudhry2019continual}& $28.2 \pm 1.2$ & $36.6 \pm 2.0$ & $12.5 \pm 0.6$ & $44.9 \pm 0.9$ & $21.2 \pm 0.6$ & $33.9 \pm 1.7$ & $15.2 \pm 0.5$ & $35.6 \pm 0.6$ \\
    GDumb\cite{prabhu2020gdumb} & $18.5 \pm 0.5$ & - & - & - & $13.1 \pm 0.4$ & - & - & - \\
    AGEM\cite{DBLP:conf/iclr/ChaudhryRRE19}& $3.1 \pm 0.2$ & $10.4 \pm 0.6$ & $2.9 \pm 0.3$ & $18.7 \pm 0.8$ & $2.6 \pm 0.2$ & $7.3 \pm 0.5$ & $2.6 \pm 0.2$ & $23.3 \pm 0.6$ \\
    $\mathrm{ER}+\mathrm{LwF}$\cite{li2017learning} & $30.4 \pm 0.8$ & $39.2 \pm 2.0$ & $15.3 \pm 0.9$ & $44.4 \pm 0.8$ & $22.7 \pm 1.1$ & $34.4 \pm 2.4$ & $17.0 \pm 0.7$ & $33.8 \pm 0.9$ \\
    MIR\cite{aljundi2019online} & $29.4 \pm 1.9$ & $33.1 \pm 3.2$ & $11.6 \pm 1.6$ & $43.4 \pm 0.7$ & $21.3 \pm 0.8$ & $31.0 \pm 1.8$ & $15.2 \pm 0.5$ & $33.0 \pm 0.4$ \\
    RAR\cite{kumari2022retrospective} & $28.2 \pm 1.4$ & $38.2 \pm 1.6$ & $14.9 \pm 0.7$ & $42.3 \pm 0.9$ & $15.7 \pm 0.9$ & $27.8 \pm 2.8$ & $10.1 \pm 0.9$ & $29.8 \pm 0.9$ \\
    DER++ \cite{buzzega2020dark}& $29.3 \pm 0.9$ & $37.5 \pm 2.5$ & $13.4 \pm 0.7$ & $44.0 \pm 0.8$ & $22.9 \pm 0.5$ & $34.2 \pm 4.0$ & $16.3 \pm 0.3$ & $31.5 \pm 0.9$ \\    
    ER-ACE \cite{DBLP:conf/iclr/CacciaAATPB22} & $29.9 \pm 0.6$ & $38.5 \pm 1.8$ & $14.9 \pm 0.9$ & $42.4 \pm 0.6$ & $\mathbf{23.6} \pm 0.7$ & $35.0 \pm 1.5$ & $16.8 \pm 0.7$ & $34.2 \pm 0.3$ \\
    SCR\cite{mai2021supervised} & $28.3 \pm 0.8$ & $42.1 \pm 2.1$ & $20.3 \pm 0.4$ & $37.0 \pm 0.3$ & $16.9 \pm 0.4$ & $30.7 \pm 1.5$ & $12.3 \pm 0.5$ & $22.5 \pm 0.4$ \\
    OnPro & $31.7 \pm 1.2$ & $36.6 \pm 2.5$ & $12.2 \pm 1.1$ & - & $17.1 \pm 1.5$ & $24.2 \pm 0.4$ & $8.00 \pm 0.8$ & - \\
    OCM   & $30.9 \pm 0.7$ & $33.3 \pm 1.9$ & $14.9 \pm0.4$ & - & $20.6 \pm 0.6$ & $24.8 \pm 1.1$  & $10.9 \pm 0.5$ & - \\
    LPR~\cite{yoo2024layerwise} & $33.3 \pm 0.6$ & $42.5 \pm 0.5$ & $19.3 \pm0.3$ & - & $23.1 \pm 0.2$ & $34.9 \pm 0.4$ & $16.2 \pm 0.2$ & - \\
    OCAR (ours) & $\mathbf{34.9} \pm 0.6$ & $\mathbf{48.2} \pm 1.2$ & $\mathbf{25.0} \pm 1.1$ & $\mathbf{46.2} \pm 0.6$ & $21.7 \pm 1.0$ & $\mathbf{38.3} \pm 1.4$ & $\mathbf{17.4} \pm 0.6$ & $\mathbf{38.3} \pm 0.6$ \\
    \midrule
    OCAR-DER++ (ours) & $34.3 \pm 1.1$& $46.8 \pm 1.7$ & $25.4 \pm 0.8$ & $46.0 \pm 0.8$ & - & - & - &- \\
    OCAR-ACE (ours) & $\underline{35.6} \pm 1.2$ & $\underline{48.7} \pm 1.7$ & $\underline{26.5} \pm 0.4$ & $44.1 \pm 0.7$& $\underline{25.6} \pm 0.4$ & $\underline{39.8} \pm 2.0$ & $\underline{21.5} \pm 0.9$ & $34.7 \pm 0.3$\\

    \bottomrule
    \end{tabular}
    }
\end{table*}

\section{Experiments}

\subsection{Comparison Between EWC, NGD, and OCAR in a Convex Setting}
First, we compare OCAR with alternative uses of the Fisher Information, that are not commonly considered for OCL, in a small-scale convex setting. In CL, EWC \cite{kirkpatrick2017overcoming} generated a cascade of derived methods~\cite{DBLP:conf/eccv/ChaudhryDAT18,liu2018rotate,DBLP:journals/corr/abs-1712-03847} based on the idea to add a quadratic regularization term to the loss, penalizing the movement of the parameters from an optimal configuration, weighted by the FIM. This approach finds some limitations in OCL when no task boundaries are provided and it's not possible to select the previous task's "best" weights \cite{mai2022online}, breaking the fundamental assumption behind the Laplace approximation. Additionally, regularizing the loss does not directly speed up non-important directions, a non-optimal approach in OCL. On the other hand, outside CL, the Natural Gradient Descent (NGD) \cite{amari1998natural} uses the FIM as a preconditioner for the gradient, slowing it down in the direction of high curvature and accelerating it in others. We believe NGD can be well-suited for OCL problems. The problem is that raw NGD is derived for i.i.d. settings. OCAR, on the other hand, is an adaptation for non-i.i.d. problems. To underline the differences, we tested the three approaches in combination with ER in an online stream of 10 small convex tasks (all details in Appendix \ref{app: qualitative}). We measure the cumulative loss experienced on single batches during the training $L_p = \sum_{t} L(y_{t}, \hat{y}_t)$ to measure the ability to adapt to current data and the cumulative loss experienced on all previous data $L_s = \sum_{t} L(y_{0:t}, \hat{y}_{0:t})$ to measure the stability of the model. The results in figure \ref{fig:qualitative} show that, while NG is much more adaptable than EWC, it is slightly less stable. EWC performance in OCL is very similar to the ones of basic ER, a result aligned with \cite{mai2022online}. OCAR, thanks to its explicit memory constraint, and its dynamic hyperparameters is able to improve both on the speed and on the stability, showing a slight optimization superiority already in this very basic setting.

\subsection{Literature benchmarks and SOTA comparison}
To ensure reproducibility and a fair comparison we use the same code and experimental setup in \cite{DBLP:conf/iccvw/Soutif-Cormerais23}, later used by  LPR \cite{yoo2024layerwise}, an ICML24 paper, that represents the SOTA in OCL and our main "competitor". We use Avalanche~\cite{carta2023avalanche} and nngeometry \cite{george_nngeometry}. In line with the literature Split-CIFAR100 (20 Task) and Split-TinyImageNet (20 tasks), are used as task incremental benchmarks. Following \cite{yoo2024layerwise} we experimented also on Online CLEAR (10 tasks) \cite{lin2021clear}, a domain incremental scenario fundamentally different from the other two datasets. 
All the methods use a reservoir sampling buffer with 2000 samples for Split-CIFAR100 and online CLEAR and 4000 for Split-TinyImageNet. More details about the experimental setting are available in the Appendix \ref{app:experimental setup} with specific information for CLEAR in Appendix \ref{app:clear}. Our entire code for the experiments can be found at \url{https://anonymous.4open.science/r/CAR-8412}.

We compare our approach with a large set of CL baselines (Table \ref{tab:tinyimnet}), evaluated on 4 fundamental OCL metrics: 
\textbf{WC-Acc} \cite{lirias4071238}: the worst-case accuracy is a metric of the stability of the model, measuring a trade-off between the model accuracy and the minimum accuracy among all tasks encountered. 
\textbf{AAA} \cite{DBLP:conf/iclr/CacciaAATPB22}: The Average Anytime Accuracy is a metric developed specifically for OCL measuring the mean accuracy of the model in time on all the encountered tasks. It is a measure of the whole accuracy history of the model.
 \textbf{Acc} \cite{DBLP:conf/nips/Lopez-PazR17}: The final average accuracy is a snapshot of the accuracy of the model at the end of the entire stream. 
\textbf{Probed Acc} \cite{davari2022probing}: The metric is the final accuracy after freezing the feature extractor and retraining only the linear classifier on all the training data. It is a measure of the representation quality of the model. 

\subsection{Results}

Using the same code and setup, all results of other methods are taken from \cite{DBLP:conf/iccvw/Soutif-Cormerais23} and \cite{yoo2024layerwise} (for LPR), except for OnPro~\cite{wei2023online} and OCM~\cite{guo2022online}, where we reused the original code and run the experiments following the setup in \cite{DBLP:conf/iccvw/Soutif-Cormerais23,yoo2024layerwise}. Everything is evaluated on 5 runs (except LPR in 10).

\textbf{End of training metrics:} While in Split-Cifar100 OCAR is able to achieve the best \textit{Acc} among base models, ER-ACE takes the crown in Tinyimagenet (Table \ref{tab:tinyimnet}). However, we must point out that \textit{Acc} is a very poor evaluation metric for the OCL performance. Since it is evaluated at the very final iteration it badly represents the whole training and can be affected by noise (check learning curves in \ref{app: figures}). OCAR obtains the best results on linear probing in both benchmarks, underlying the optimization improvements also on the feature extractor. This is a significant result when compared to the i.i.d. case: while the classifier suffers from forgetting, OCAR can learn a better feature representation if compared to the i.i.d. case, confirming how it learns efficiently in nonstationary settings.

\textbf{Continual metrics:} When evaluated at every step in time, OCAR obtains the best \textit{AAA} and \textit{WC-Acc} among the methods in both benchmarks. This confirms OCAR as a robust continual optimizer with high accuracy and stability at every point in time. Both metrics are significantly improved with a jump of several points. 

\textbf{Integration with other methods:} Being ER-ACE the method with the best accuracy on Tinyimagenet, we try a combination with it to see if OCAR can improve the results. Even if ER-ACE uses a slightly modified loss, the integration with OCAR works very well. The combination OCAR-ACE beats all other methods (including base OCAR) on \textit{Acc}, \textit{AAA} and \textit{WC-Acc}. Both ER-ACE and OCAR-ACE show a slightly lower probing accuracy, which suggests that ER-ACE may be tuned to prefer stability over plasticity, learning less transferable features despite the high accuracy. Possibly, the robustness of the classifier given by ER-ACE slightly interferes with deeper feature learning. Following this experiment, we also tried the combination with DER++. Unfortunately, we have found OCAR-DER to be much less stable (failing optimization on Tinyimagent), which we conjecture to be caused by the entropy regularization implicit in the DER loss, which results in a loss that is "too different" from the KL divergence and breaks the assumption for the use of the Fisher.

\textbf{Online CLEAR}
The Online CLEAR benchmark is fundamentally different: no classes are added from task to task, but all classes evolve in time, in a domain incremental fashion. This puts much more importance on forward and backward transfer and less on catastrophic forgetting due to the similarities between tasks. For this reason, the final accuracy is higher than \textit{AAA} (see figure \ref{fig:clear accuracy} in Appendix). This setting, being more similar to the standard iid one, makes much easier the estimation and the stability of the FIM, making OCAR remarkably better than the previous SOTA LPR (see table \ref{tab:clear}). This confirms OCAR robustness also on domain incremental settings, underlying its optimization improvements. 
\begin{table}[h!]
    \caption{Results on Online CLEAR (10 Tasks) domain incremental setting. 2000 Buffer size. Best in bold.}\label{tab:clear}
    {%
    \begin{tabular}{l|ccc}
    \hline \multirow[t]{2}{*}{Method} & \multicolumn{3}{c}{Online CLEAR (10 Tasks)}  \\
    & Acc $\uparrow$ & $A A A^{\text {val }} \uparrow$ & WC-Acc ${ }^{\text {val }} \uparrow$ \\
    \midrule
    ER & $63.1 \pm 0.7$ & $58.9 \pm 0.8$ & $47.7 \pm 1.6$ \\ 
    LPR & $65.2 \pm 0.9$ & $63.5 \pm 1.0$ & $62.6 \pm 0.7$ \\
    \midrule
    OCAR(Ours) & $\mathbf{75.3} \pm 0.8$ & $\mathbf{73.9} \pm 0.5$ & $\mathbf{70.3} \pm 0.5$\\
    \bottomrule
    \end{tabular}
    }
\end{table}

\textbf{Final Comment}: OCAR showed remarkable performance across all continual metrics, improving both on task-incremental and domain-incremental the previous SOTA results, including the ICML24 paper LPR \cite{yoo2024layerwise}, using the same code, setting and benchmarks. All of this is done efficiently, even improving on computational time (see Appendix \ref{app:experimental setup}) of some previous SOTA approaches. The method showed the possibility of being combined with other approaches, obtaining even stronger results.



\section{Conclusion}
In this paper, we revisit replay-based OCL as second-order optimization with hard stability constraints and information geometry rooting. The resulting method OCAR shows consistent improvements in plasticity and continual stability with clear hyperparameters interpretation in the stability-plasticity tradeoff.
Future research directions include the comparison of different approximations of the curvature (e.g. \citet{george2018fast}), alternative derivations for the optimization problem (e.g. \citet{benzing2022gradient}), and feasible dynamic adaptation of $\alpha$ and $\tau$. We believe OCAR can be the starting point for further improvements towards a deeper understanding of continual learning dynamics.

\section*{Impact Statement}
This paper presents work whose goal is to advance the field of Machine Learning. There are many potential societal consequences of our work, none which we feel must be specifically highlighted here.


\bibliography{references}

\begin{thebibliography}{59}
\providecommand{\natexlab}[1]{#1}
\providecommand{\url}[1]{\texttt{#1}}
\expandafter\ifx\csname urlstyle\endcsname\relax
  \providecommand{\doi}[1]{doi: #1}\else
  \providecommand{\doi}{doi: \begingroup \urlstyle{rm}\Url}\fi

\bibitem[Aljundi et~al.(2019{\natexlab{a}})Aljundi, Belilovsky, Tuytelaars, Charlin, Caccia, Lin, and Page-Caccia]{aljundi2019online}
Aljundi, R., Belilovsky, E., Tuytelaars, T., Charlin, L., Caccia, M., Lin, M., and Page-Caccia, L.
\newblock Online continual learning with maximal interfered retrieval.
\newblock \emph{Advances in neural information processing systems}, 32, 2019{\natexlab{a}}.

\bibitem[Aljundi et~al.(2019{\natexlab{b}})Aljundi, Kelchtermans, and Tuytelaars]{aljundi2019task}
Aljundi, R., Kelchtermans, K., and Tuytelaars, T.
\newblock Task-free continual learning.
\newblock In \emph{Proceedings of the IEEE/CVF conference on computer vision and pattern recognition}, pp.\  11254--11263, 2019{\natexlab{b}}.

\bibitem[Amari(1998)]{amari1998natural}
Amari, S.-I.
\newblock Natural gradient works efficiently in learning.
\newblock \emph{Neural computation}, 10\penalty0 (2):\penalty0 251--276, 1998.

\bibitem[Amari(2016)]{amari2016information}
Amari, S.-i.
\newblock \emph{Information geometry and its applications}, volume 194.
\newblock Springer, 2016.

\bibitem[Benzing(2022{\natexlab{a}})]{DBLP:conf/icml/Benzing22}
Benzing, F.
\newblock Gradient descent on neurons and its link to approximate second-order optimization.
\newblock In Chaudhuri, K., Jegelka, S., Song, L., Szepesv{\'{a}}ri, C., Niu, G., and Sabato, S. (eds.), \emph{International Conference on Machine Learning, {ICML} 2022, 17-23 July 2022, Baltimore, Maryland, {USA}}, volume 162 of \emph{Proceedings of Machine Learning Research}, pp.\  1817--1853. {PMLR}, 2022{\natexlab{a}}.
\newblock URL \url{https://proceedings.mlr.press/v162/benzing22a.html}.

\bibitem[Benzing(2022{\natexlab{b}})]{benzing2022gradient}
Benzing, F.
\newblock Gradient descent on neurons and its link to approximate second-order optimization.
\newblock In \emph{International Conference on Machine Learning}, pp.\  1817--1853. PMLR, 2022{\natexlab{b}}.

\bibitem[Buzzega et~al.(2020)Buzzega, Boschini, Porrello, Abati, and Calderara]{buzzega2020dark}
Buzzega, P., Boschini, M., Porrello, A., Abati, D., and Calderara, S.
\newblock Dark experience for general continual learning: a strong, simple baseline.
\newblock \emph{Advances in neural information processing systems}, 33:\penalty0 15920--15930, 2020.

\bibitem[Caccia et~al.(2022)Caccia, Aljundi, Asadi, Tuytelaars, Pineau, and Belilovsky]{DBLP:conf/iclr/CacciaAATPB22}
Caccia, L., Aljundi, R., Asadi, N., Tuytelaars, T., Pineau, J., and Belilovsky, E.
\newblock New insights on reducing abrupt representation change in online continual learning.
\newblock In \emph{The Tenth International Conference on Learning Representations, {ICLR} 2022, Virtual Event, April 25-29, 2022}. OpenReview.net, 2022.
\newblock URL \url{https://openreview.net/forum?id=N8MaByOzUfb}.

\bibitem[Carta et~al.(2023)Carta, Pellegrini, Cossu, Hemati, and Lomonaco]{carta2023avalanche}
Carta, A., Pellegrini, L., Cossu, A., Hemati, H., and Lomonaco, V.
\newblock Avalanche: A pytorch library for deep continual learning.
\newblock \emph{Journal of Machine Learning Research}, 24\penalty0 (363):\penalty0 1--6, 2023.

\bibitem[Chaudhry et~al.(2018{\natexlab{a}})Chaudhry, Dokania, Ajanthan, and Torr]{chaudhry2018riemannian}
Chaudhry, A., Dokania, P.~K., Ajanthan, T., and Torr, P.~H.
\newblock Riemannian walk for incremental learning: Understanding forgetting and intransigence.
\newblock In \emph{Proceedings of the European conference on computer vision (ECCV)}, pp.\  532--547, 2018{\natexlab{a}}.

\bibitem[Chaudhry et~al.(2018{\natexlab{b}})Chaudhry, Dokania, Ajanthan, and Torr]{DBLP:conf/eccv/ChaudhryDAT18}
Chaudhry, A., Dokania, P.~K., Ajanthan, T., and Torr, P. H.~S.
\newblock Riemannian walk for incremental learning: Understanding forgetting and intransigence.
\newblock In Ferrari, V., Hebert, M., Sminchisescu, C., and Weiss, Y. (eds.), \emph{Computer Vision - {ECCV} 2018 - 15th European Conference, Munich, Germany, September 8-14, 2018, Proceedings, Part {XI}}, volume 11215 of \emph{Lecture Notes in Computer Science}, pp.\  556--572. Springer, 2018{\natexlab{b}}.
\newblock \doi{10.1007/978-3-030-01252-6\_33}.
\newblock URL \url{https://doi.org/10.1007/978-3-030-01252-6\_33}.

\bibitem[Chaudhry et~al.(2019{\natexlab{a}})Chaudhry, Ranzato, Rohrbach, and Elhoseiny]{DBLP:conf/iclr/ChaudhryRRE19}
Chaudhry, A., Ranzato, M., Rohrbach, M., and Elhoseiny, M.
\newblock Efficient lifelong learning with {A-GEM}.
\newblock In \emph{7th International Conference on Learning Representations, {ICLR} 2019, New Orleans, LA, USA, May 6-9, 2019}. OpenReview.net, 2019{\natexlab{a}}.
\newblock URL \url{https://openreview.net/forum?id=Hkf2\_sC5FX}.

\bibitem[Chaudhry et~al.(2019{\natexlab{b}})Chaudhry, Rohrbach, Elhoseiny, Ajanthan, Dokania, Torr, and Ranzato]{chaudhry2019continual}
Chaudhry, A., Rohrbach, M., Elhoseiny, M., Ajanthan, T., Dokania, P., Torr, P., and Ranzato, M.
\newblock Continual learning with tiny episodic memories.
\newblock In \emph{Workshop on Multi-Task and Lifelong Reinforcement Learning}, 2019{\natexlab{b}}.

\bibitem[Davari et~al.(2022)Davari, Asadi, Mudur, Aljundi, and Belilovsky]{davari2022probing}
Davari, M., Asadi, N., Mudur, S., Aljundi, R., and Belilovsky, E.
\newblock Probing representation forgetting in supervised and unsupervised continual learning.
\newblock In \emph{Proceedings of the IEEE/CVF Conference on Computer Vision and Pattern Recognition}, pp.\  16712--16721, 2022.

\bibitem[Daxberger et~al.(2023)Daxberger, Swaroop, Osawa, Yokota, Turner, Hern{\'a}ndez-Lobato, and Khan]{daxberger2023improving}
Daxberger, E., Swaroop, S., Osawa, K., Yokota, R., Turner, R.~E., Hern{\'a}ndez-Lobato, J.~M., and Khan, M.~E.
\newblock Improving continual learning by accurate gradient reconstructions of the past.
\newblock \emph{Transactions on Machine Learning Research}, 2023.

\bibitem[De~Lange et~al.(2023)De~Lange, van~de Ven, and Tuytelaars]{lirias4071238}
De~Lange, M., van~de Ven, G., and Tuytelaars, T.
\newblock Continual evaluation for lifelong learning: Identifying the stability gap, 2023.

\bibitem[Dohare et~al.(2024)Dohare, Hernandez{-}Garcia, Lan, Rahman, Mahmood, and Sutton]{DBLP:journals/nature/DohareHLRMS24}
Dohare, S., Hernandez{-}Garcia, J.~F., Lan, Q., Rahman, P., Mahmood, A.~R., and Sutton, R.~S.
\newblock Loss of plasticity in deep continual learning.
\newblock \emph{Nat.}, 632\penalty0 (8026):\penalty0 768--774, 2024.
\newblock \doi{10.1038/S41586-024-07711-7}.
\newblock URL \url{https://doi.org/10.1038/s41586-024-07711-7}.

\bibitem[Durbin \& Koopman(2012)Durbin and Koopman]{durbin2012time}
Durbin, J. and Koopman, S.~J.
\newblock \emph{Time series analysis by state space methods}, volume~38.
\newblock OUP Oxford, 2012.

\bibitem[Farajtabar et~al.(2020)Farajtabar, Azizan, Mott, and Li]{DBLP:conf/aistats/FarajtabarAML20}
Farajtabar, M., Azizan, N., Mott, A., and Li, A.
\newblock Orthogonal gradient descent for continual learning.
\newblock In Chiappa, S. and Calandra, R. (eds.), \emph{The 23rd International Conference on Artificial Intelligence and Statistics, {AISTATS} 2020, 26-28 August 2020, Online [Palermo, Sicily, Italy]}, volume 108 of \emph{Proceedings of Machine Learning Research}, pp.\  3762--3773. {PMLR}, 2020.
\newblock URL \url{http://proceedings.mlr.press/v108/farajtabar20a.html}.

\bibitem[French(1999)]{french1999catastrophic}
French, R.~M.
\newblock Catastrophic forgetting in connectionist networks.
\newblock \emph{Trends in cognitive sciences}, 3\penalty0 (4):\penalty0 128--135, 1999.

\bibitem[George(2021)]{george_nngeometry}
George, T.
\newblock {NNGeometry: Easy and Fast Fisher Information Matrices and Neural Tangent Kernels in PyTorch}, February 2021.
\newblock URL \url{https://doi.org/10.5281/zenodo.4532597}.

\bibitem[George et~al.(2018{\natexlab{a}})George, Laurent, Bouthillier, Ballas, and Vincent]{DBLP:conf/nips/GeorgeLBBV18}
George, T., Laurent, C., Bouthillier, X., Ballas, N., and Vincent, P.
\newblock Fast approximate natural gradient descent in a kronecker factored eigenbasis.
\newblock In Bengio, S., Wallach, H.~M., Larochelle, H., Grauman, K., Cesa{-}Bianchi, N., and Garnett, R. (eds.), \emph{Advances in Neural Information Processing Systems 31: Annual Conference on Neural Information Processing Systems 2018, NeurIPS 2018, December 3-8, 2018, Montr{\'{e}}al, Canada}, pp.\  9573--9583, 2018{\natexlab{a}}.
\newblock URL \url{https://proceedings.neurips.cc/paper/2018/hash/48000647b315f6f00f913caa757a70b3-Abstract.html}.

\bibitem[George et~al.(2018{\natexlab{b}})George, Laurent, Bouthillier, Ballas, and Vincent]{george2018fast}
George, T., Laurent, C., Bouthillier, X., Ballas, N., and Vincent, P.
\newblock Fast approximate natural gradient descent in a kronecker factored eigenbasis.
\newblock \emph{Advances in Neural Information Processing Systems}, 31, 2018{\natexlab{b}}.

\bibitem[Guo et~al.(2022)Guo, Liu, and Zhao]{guo2022online}
Guo, Y., Liu, B., and Zhao, D.
\newblock Online continual learning through mutual information maximization.
\newblock In \emph{International conference on machine learning}, pp.\  8109--8126. PMLR, 2022.

\bibitem[Hess et~al.(2023)Hess, Tuytelaars, and van~de Ven]{DBLP:journals/corr/abs-2311-04898}
Hess, T., Tuytelaars, T., and van~de Ven, G.~M.
\newblock Two complementary perspectives to continual learning: Ask not only what to optimize, but also how.
\newblock \emph{CoRR}, abs/2311.04898, 2023.
\newblock \doi{10.48550/ARXIV.2311.04898}.
\newblock URL \url{https://doi.org/10.48550/arXiv.2311.04898}.

\bibitem[Husz{\'{a}}r(2017)]{DBLP:journals/corr/abs-1712-03847}
Husz{\'{a}}r, F.
\newblock On quadratic penalties in elastic weight consolidation.
\newblock \emph{CoRR}, abs/1712.03847, 2017.
\newblock URL \url{http://arxiv.org/abs/1712.03847}.

\bibitem[Kamath et~al.(2024)Kamath, Soutif{-}Cormerais, van~de Weijer, and Raducanu]{DBLP:journals/corr/abs-2406-05114}
Kamath, S., Soutif{-}Cormerais, A., van~de Weijer, J., and Raducanu, B.
\newblock The expanding scope of the stability gap: Unveiling its presence in joint incremental learning of homogeneous tasks.
\newblock \emph{CoRR}, abs/2406.05114, 2024.
\newblock \doi{10.48550/ARXIV.2406.05114}.
\newblock URL \url{https://doi.org/10.48550/arXiv.2406.05114}.

\bibitem[Kao et~al.(2021)Kao, Jensen, van~de Ven, Bernacchia, and Hennequin]{DBLP:conf/nips/KaoJVBH21}
Kao, T., Jensen, K.~T., van~de Ven, G., Bernacchia, A., and Hennequin, G.
\newblock Natural continual learning: success is a journey, not (just) a destination.
\newblock In Ranzato, M., Beygelzimer, A., Dauphin, Y.~N., Liang, P., and Vaughan, J.~W. (eds.), \emph{Advances in Neural Information Processing Systems 34: Annual Conference on Neural Information Processing Systems 2021, NeurIPS 2021, December 6-14, 2021, virtual}, pp.\  28067--28079, 2021.
\newblock URL \url{https://proceedings.neurips.cc/paper/2021/hash/ec5aa0b7846082a2415f0902f0da88f2-Abstract.html}.

\bibitem[Kingma(2014)]{kingma2014adam}
Kingma, D.~P.
\newblock Adam: A method for stochastic optimization.
\newblock \emph{arXiv preprint arXiv:1412.6980}, 2014.

\bibitem[Kingma \& Ba(2015)Kingma and Ba]{DBLP:journals/corr/KingmaB14}
Kingma, D.~P. and Ba, J.
\newblock Adam: {A} method for stochastic optimization.
\newblock In Bengio, Y. and LeCun, Y. (eds.), \emph{3rd International Conference on Learning Representations, {ICLR} 2015, San Diego, CA, USA, May 7-9, 2015, Conference Track Proceedings}, 2015.
\newblock URL \url{http://arxiv.org/abs/1412.6980}.

\bibitem[Kirkpatrick et~al.(2017)Kirkpatrick, Pascanu, Rabinowitz, Veness, Desjardins, Rusu, Milan, Quan, Ramalho, Grabska-Barwinska, et~al.]{kirkpatrick2017overcoming}
Kirkpatrick, J., Pascanu, R., Rabinowitz, N., Veness, J., Desjardins, G., Rusu, A.~A., Milan, K., Quan, J., Ramalho, T., Grabska-Barwinska, A., et~al.
\newblock Overcoming catastrophic forgetting in neural networks.
\newblock \emph{Proceedings of the national academy of sciences}, 114\penalty0 (13):\penalty0 3521--3526, 2017.

\bibitem[Kumari et~al.(2022)Kumari, Wang, Zhou, and Bilmes]{kumari2022retrospective}
Kumari, L., Wang, S., Zhou, T., and Bilmes, J.~A.
\newblock Retrospective adversarial replay for continual learning.
\newblock \emph{Advances in neural information processing systems}, 35:\penalty0 28530--28544, 2022.

\bibitem[Kunstner et~al.(2019)Kunstner, Hennig, and Balles]{kunstner2019limitations}
Kunstner, F., Hennig, P., and Balles, L.
\newblock Limitations of the empirical fisher approximation for natural gradient descent.
\newblock \emph{Advances in neural information processing systems}, 32, 2019.

\bibitem[Lange et~al.(2022)Lange, Aljundi, Masana, Parisot, Jia, Leonardis, Slabaugh, and Tuytelaars]{DBLP:journals/pami/LangeAMPJLST22}
Lange, M.~D., Aljundi, R., Masana, M., Parisot, S., Jia, X., Leonardis, A., Slabaugh, G.~G., and Tuytelaars, T.
\newblock A continual learning survey: Defying forgetting in classification tasks.
\newblock \emph{{IEEE} Trans. Pattern Anal. Mach. Intell.}, 44\penalty0 (7):\penalty0 3366--3385, 2022.
\newblock \doi{10.1109/TPAMI.2021.3057446}.
\newblock URL \url{https://doi.org/10.1109/TPAMI.2021.3057446}.

\bibitem[Lange et~al.(2023)Lange, van~de Ven, and Tuytelaars]{DBLP:conf/iclr/LangeVT23}
Lange, M.~D., van~de Ven, G.~M., and Tuytelaars, T.
\newblock Continual evaluation for lifelong learning: Identifying the stability gap.
\newblock In \emph{The Eleventh International Conference on Learning Representations, {ICLR} 2023, Kigali, Rwanda, May 1-5, 2023}. OpenReview.net, 2023.
\newblock URL \url{https://openreview.net/forum?id=Zy350cRstc6}.

\bibitem[LeCun et~al.(2015)LeCun, Bengio, and Hinton]{lecun2015deep}
LeCun, Y., Bengio, Y., and Hinton, G.
\newblock Deep learning.
\newblock \emph{nature}, 521\penalty0 (7553):\penalty0 436--444, 2015.

\bibitem[Li \& Hoiem(2017)Li and Hoiem]{li2017learning}
Li, Z. and Hoiem, D.
\newblock Learning without forgetting.
\newblock \emph{IEEE transactions on pattern analysis and machine intelligence}, 40\penalty0 (12):\penalty0 2935--2947, 2017.

\bibitem[Lin et~al.(2021)Lin, Shi, Pathak, and Ramanan]{lin2021clear}
Lin, Z., Shi, J., Pathak, D., and Ramanan, D.
\newblock The clear benchmark: Continual learning on real-world imagery.
\newblock In \emph{Thirty-fifth conference on neural information processing systems datasets and benchmarks track (round 2)}, 2021.

\bibitem[Liu et~al.(2018)Liu, Masana, Herranz, Van~de Weijer, Lopez, and Bagdanov]{liu2018rotate}
Liu, X., Masana, M., Herranz, L., Van~de Weijer, J., Lopez, A.~M., and Bagdanov, A.~D.
\newblock Rotate your networks: Better weight consolidation and less catastrophic forgetting.
\newblock In \emph{2018 24th International Conference on Pattern Recognition (ICPR)}, pp.\  2262--2268. IEEE, 2018.

\bibitem[Lopez{-}Paz \& Ranzato(2017)Lopez{-}Paz and Ranzato]{DBLP:conf/nips/Lopez-PazR17}
Lopez{-}Paz, D. and Ranzato, M.
\newblock Gradient episodic memory for continual learning.
\newblock In Guyon, I., von Luxburg, U., Bengio, S., Wallach, H.~M., Fergus, R., Vishwanathan, S. V.~N., and Garnett, R. (eds.), \emph{Advances in Neural Information Processing Systems 30: Annual Conference on Neural Information Processing Systems 2017, December 4-9, 2017, Long Beach, CA, {USA}}, pp.\  6467--6476, 2017.
\newblock URL \url{https://proceedings.neurips.cc/paper/2017/hash/f87522788a2be2d171666752f97ddebb-Abstract.html}.

\bibitem[Magistri et~al.(2024)Magistri, Trinci, Soutif{-}Cormerais, van~de Weijer, and Bagdanov]{DBLP:conf/iclr/MagistriTS0B24}
Magistri, S., Trinci, T., Soutif{-}Cormerais, A., van~de Weijer, J., and Bagdanov, A.~D.
\newblock Elastic feature consolidation for cold start exemplar-free incremental learning.
\newblock In \emph{The Twelfth International Conference on Learning Representations, {ICLR} 2024, Vienna, Austria, May 7-11, 2024}. OpenReview.net, 2024.
\newblock URL \url{https://openreview.net/forum?id=7D9X2cFnt1}.

\bibitem[Mai et~al.(2021)Mai, Li, Kim, and Sanner]{mai2021supervised}
Mai, Z., Li, R., Kim, H., and Sanner, S.
\newblock Supervised contrastive replay: Revisiting the nearest class mean classifier in online class-incremental continual learning.
\newblock In \emph{Proceedings of the IEEE/CVF conference on computer vision and pattern recognition}, pp.\  3589--3599, 2021.

\bibitem[Mai et~al.(2022{\natexlab{a}})Mai, Li, Jeong, Quispe, Kim, and Sanner]{DBLP:journals/ijon/MaiLJQKS22}
Mai, Z., Li, R., Jeong, J., Quispe, D., Kim, H., and Sanner, S.
\newblock Online continual learning in image classification: An empirical survey.
\newblock \emph{Neurocomputing}, 469:\penalty0 28--51, 2022{\natexlab{a}}.
\newblock \doi{10.1016/J.NEUCOM.2021.10.021}.
\newblock URL \url{https://doi.org/10.1016/j.neucom.2021.10.021}.

\bibitem[Mai et~al.(2022{\natexlab{b}})Mai, Li, Jeong, Quispe, Kim, and Sanner]{mai2022online}
Mai, Z., Li, R., Jeong, J., Quispe, D., Kim, H., and Sanner, S.
\newblock Online continual learning in image classification: An empirical survey.
\newblock \emph{Neurocomputing}, 469:\penalty0 28--51, 2022{\natexlab{b}}.

\bibitem[Martens(2020)]{martens2020new}
Martens, J.
\newblock New insights and perspectives on the natural gradient method.
\newblock \emph{Journal of Machine Learning Research}, 21\penalty0 (146):\penalty0 1--76, 2020.

\bibitem[Martens \& Grosse(2015)Martens and Grosse]{martens2015optimizing}
Martens, J. and Grosse, R.
\newblock Optimizing neural networks with kronecker-factored approximate curvature.
\newblock In \emph{International conference on machine learning}, pp.\  2408--2417. PMLR, 2015.

\bibitem[Martens \& Sutskever(2012)Martens and Sutskever]{martens2012training}
Martens, J. and Sutskever, I.
\newblock Training deep and recurrent networks with hessian-free optimization.
\newblock In \emph{Neural Networks: Tricks of the Trade: Second Edition}, pp.\  479--535. Springer, 2012.

\bibitem[Masana et~al.(2023)Masana, Liu, Twardowski, Menta, Bagdanov, and van~de Weijer]{DBLP:journals/pami/MasanaLTMBW23}
Masana, M., Liu, X., Twardowski, B., Menta, M., Bagdanov, A.~D., and van~de Weijer, J.
\newblock Class-incremental learning: Survey and performance evaluation on image classification.
\newblock \emph{{IEEE} Trans. Pattern Anal. Mach. Intell.}, 45\penalty0 (5):\penalty0 5513--5533, 2023.
\newblock \doi{10.1109/TPAMI.2022.3213473}.
\newblock URL \url{https://doi.org/10.1109/TPAMI.2022.3213473}.

\bibitem[Mirzadeh et~al.(2020)Mirzadeh, Farajtabar, Pascanu, and Ghasemzadeh]{DBLP:conf/nips/MirzadehFPG20}
Mirzadeh, S., Farajtabar, M., Pascanu, R., and Ghasemzadeh, H.
\newblock Understanding the role of training regimes in continual learning.
\newblock In Larochelle, H., Ranzato, M., Hadsell, R., Balcan, M., and Lin, H. (eds.), \emph{Advances in Neural Information Processing Systems 33: Annual Conference on Neural Information Processing Systems 2020, NeurIPS 2020, December 6-12, 2020, virtual}, 2020.
\newblock URL \url{https://proceedings.neurips.cc/paper/2020/hash/518a38cc9a0173d0b2dc088166981cf8-Abstract.html}.

\bibitem[Mirzadeh et~al.(2021)Mirzadeh, Farajtabar, G{\"{o}}r{\"{u}}r, Pascanu, and Ghasemzadeh]{DBLP:conf/iclr/MirzadehFGP021}
Mirzadeh, S., Farajtabar, M., G{\"{o}}r{\"{u}}r, D., Pascanu, R., and Ghasemzadeh, H.
\newblock Linear mode connectivity in multitask and continual learning.
\newblock In \emph{9th International Conference on Learning Representations, {ICLR} 2021, Virtual Event, Austria, May 3-7, 2021}. OpenReview.net, 2021.
\newblock URL \url{https://openreview.net/forum?id=Fmg\_fQYUejf}.

\bibitem[Ollivier et~al.(2017)Ollivier, Arnold, Auger, and Hansen]{ollivier2017information}
Ollivier, Y., Arnold, L., Auger, A., and Hansen, N.
\newblock Information-geometric optimization algorithms: A unifying picture via invariance principles.
\newblock \emph{Journal of Machine Learning Research}, 18\penalty0 (18):\penalty0 1--65, 2017.

\bibitem[Pan et~al.(2020)Pan, Swaroop, Immer, Eschenhagen, Turner, and Khan]{pan2020continual}
Pan, P., Swaroop, S., Immer, A., Eschenhagen, R., Turner, R., and Khan, M. E.~E.
\newblock Continual deep learning by functional regularisation of memorable past.
\newblock \emph{Advances in neural information processing systems}, 33:\penalty0 4453--4464, 2020.

\bibitem[Prabhu et~al.(2020)Prabhu, Torr, and Dokania]{prabhu2020gdumb}
Prabhu, A., Torr, P.~H., and Dokania, P.~K.
\newblock Gdumb: A simple approach that questions our progress in continual learning.
\newblock In \emph{Computer Vision--ECCV 2020: 16th European Conference, Glasgow, UK, August 23--28, 2020, Proceedings, Part II 16}, pp.\  524--540. Springer, 2020.

\bibitem[Saha et~al.(2021)Saha, Garg, and Roy]{DBLP:conf/iclr/SahaG021}
Saha, G., Garg, I., and Roy, K.
\newblock Gradient projection memory for continual learning.
\newblock In \emph{9th International Conference on Learning Representations, {ICLR} 2021, Virtual Event, Austria, May 3-7, 2021}. OpenReview.net, 2021.
\newblock URL \url{https://openreview.net/forum?id=3AOj0RCNC2}.

\bibitem[Soutif{-}Cormerais et~al.(2023)Soutif{-}Cormerais, Carta, Cossu, Hurtado, Lomonaco, van~de Weijer, and Hemati]{DBLP:conf/iccvw/Soutif-Cormerais23}
Soutif{-}Cormerais, A., Carta, A., Cossu, A., Hurtado, J., Lomonaco, V., van~de Weijer, J., and Hemati, H.
\newblock A comprehensive empirical evaluation on online continual learning.
\newblock In \emph{{IEEE/CVF} International Conference on Computer Vision, {ICCV} 2023 - Workshops, Paris, France, October 2-6, 2023}, pp.\  3510--3520. {IEEE}, 2023.
\newblock \doi{10.1109/ICCVW60793.2023.00378}.
\newblock URL \url{https://doi.org/10.1109/ICCVW60793.2023.00378}.

\bibitem[Thomas \& Joy(2006)Thomas and Joy]{thomas2006elements}
Thomas, M. and Joy, A.~T.
\newblock \emph{Elements of information theory}.
\newblock Wiley-Interscience, 2006.

\bibitem[van~de Ven \& Tolias(2019)van~de Ven and Tolias]{DBLP:journals/corr/abs-1904-07734}
van~de Ven, G.~M. and Tolias, A.~S.
\newblock Three scenarios for continual learning.
\newblock \emph{CoRR}, abs/1904.07734, 2019.
\newblock URL \url{http://arxiv.org/abs/1904.07734}.

\bibitem[Wei et~al.(2023)Wei, Ye, Huang, Zhang, and Shan]{wei2023online}
Wei, Y., Ye, J., Huang, Z., Zhang, J., and Shan, H.
\newblock Online prototype learning for online continual learning.
\newblock In \emph{Proceedings of the IEEE/CVF International Conference on Computer Vision}, pp.\  18764--18774, 2023.

\bibitem[Yoo et~al.(2024)Yoo, Liu, Wood, and Pleiss]{yoo2024layerwise}
Yoo, J., Liu, Y., Wood, F., and Pleiss, G.
\newblock Layerwise proximal replay: A proximal point method for online continual learning.
\newblock \emph{arXiv preprint arXiv:2402.09542}, 2024.

\end{thebibliography}
\bibliographystyle{icml2025}

\newpage
\appendix
\onecolumn
\section{Algorithm}
\label{app: algo}
\begin{algorithm}
\caption{Online Curvature-Aware Replay (OCAR)}
\textbf{Input:} network parameters $w$, learning rate $\alpha$, per batch gradient steps count $S$, Tikhonov increase $\Delta \tau$, EMA parameter $\alpha_{EMA}$.\\
\textbf{Output:} trained network parameters $w$.
\begin{algorithmic}[1]
    \STATE Initialize replay buffer: $\mathcal{B} \gets \{\}$.
    \STATE $\tau \gets \alpha$
    \STATE $\lambda \gets 1$
    \FOR{$t \in \{1, ..., \infty\}$}
        \STATE Obtain new data batch $N_t$.
        \STATE Sample buffer data batch $B_t$
        \FOR{$s \in \{1, ..., S\}$}
            \STATE Compute loss $\mathcal{L}(w)$ using $N_t$ and $B_t$.
            \STATE Compute loss gradient $\nabla \mathcal{L}(w)$.
            \STATE $\tau \gets \tau + \Delta \tau$
            \IF{Class Incremental}
                \STATE Update known classes list with $N_t$
                \STATE Increase $\lambda$ with new classes
            \ELSE
                \STATE $\lambda \gets \lambda + \Delta\tau$
            \ENDIF
            \IF{s = 1}
                \STATE Compute K-FAC factors $A$ and $G$ with $N_t$ and $B_t$ ($B_t$ influence weighted by $\lambda$)
                \FOR{$l \in {1, ..., L}$}
                    \STATE $A_{EMA,l} \gets (1-\alpha_{EMA}) A_{EMA,l} + \alpha_{EMA} A_l$
                    \STATE $G_{EMA,l} \gets (1-\alpha_{EMA}) G_{EMA,l} + \alpha_{EMA} G_l$
                    \IF{$l = L$ and $L$ changed shape}
                    \STATE $G_{EMA,l} \gets  G_l$
                    \ENDIF
                \ENDFOR
                \STATE $F_{EMA} \gets$ ${A_{EMA}, G_{EMA}}$
                \STATE $F_{INV} \gets (F_{EMA}+\tau\mathbf{I})^{-1}$
            \ENDIF
            \STATE$ \Tilde{\nabla} \mathcal{L}(w) \gets F_{INV} \nabla \mathcal{L}(w)$
             \STATE $w \gets w - \alpha \Tilde{\nabla} \mathcal{L}(w)$
        \ENDFOR
        \STATE $B \gets$ Reservoir.update$(B, Nt, maxsize)$
    \ENDFOR
\end{algorithmic}
\end{algorithm}

The K-FAC computations are done after weighting with $\lambda$ the data related to the buffer. 

\section{Experimental Setup}
\label{app:experimental setup}
\paragraph{Code}The entire code used to perform the full experiments (CIFAR100, TinyImageNet e Clear) can be found in the anonymous repository at \url{https://anonymous.4open.science/r/CAR-8412}. The anonymization process could have removed some important parts of the code but this is highly unlikely. The code was built on a combination of the repository for the OCL survey \cite{DBLP:conf/iccvw/Soutif-Cormerais23} and the nngeometry repository \cite{george_nngeometry}.

\paragraph{Hardware}The experiments were performed in a Linux cluster equipped with Nvidia Tesla V100 16GB GPUs and Intel Xeon Gold 6140M CPUs. The training was not parallelized between the GPUs. 

\paragraph{Main training setup} All the experiments have been performed as in \cite{DBLP:conf/iccvw/Soutif-Cormerais23}. We only report the most important aspects here, referring to the original paper for all the additional details. The datasets used for the main experiments are Split-CIFAR100 and Split-TinyImageNet. Both are class-incremental benchmarks, divided in 20 tasks, with a number of new classes encountered at each new experience (5 in Split-CIFAR100 and 10 in Split-TinyImageNet). The model used for the main experiments is a Slim-ResNet18 trained using SGD. SGD can help the model adapt faster when new data are encountered after a task boundary, while ADAM \cite{kingma2014adam} will need some time to update its running statistics. The batches are tiny, with 10 examples from the current portion of the stream and 10 from the replay buffer. The replay buffer for Split-CIFAR100 was kept at 2000 samples, while for Split-TinyImageNet at 4000. The metrics computed at the end of the training (cumulative accuracy and linear probing accuracy) are computed on a test stream. The continual metrics (average anytime accuracy and worst-case accuracy) are computed after the training on each batch on a heald-out validation set. The only additional information OCAR has access to is the number of classes on each experience (5 for one and 10 for the other). This information is only used to increase the weight $\lambda$ for the computation of the Fisher Information on the buffer data. If the buffer data contains $n$ classes and the number of classes observed in the current portion of the stream is $k$, then the FIM of the buffer will weight $\frac{n}{k}$. While this information is useful, if it is not accessible, other weighting procedures can be performed, or, it can simply be estimated by watching how many classes are observed in the first few steps of the training. Usually few steps are enough to observe all the different classes present in the task. 

The same setup has been used in \cite{yoo2024layerwise}, with the only difference they used 10 seeds for their experiments instead of only 5. There is also a possible difference in the hyperparameter selection. Given the equality of the training and validation setup, we compared our methods' performances directly with the results of \cite{yoo2024layerwise} and \cite{DBLP:conf/iccvw/Soutif-Cormerais23}. This minimizes the probability of errors or bugs and uses the results obtained in the best conditions. 

\paragraph{Training time}
We report the training time to complete a single task on Split-CIFAR100 of different methods. Since OCAR and other OCL methods do not increase their computational cost over time, we report the training times on the first task to avoid unnecessary retraining of all the methods. We train on the first task without the continual evaluation to remove the evaluation overhead and provide a fair comparison between the methods. 

\begin{table}[ht]
    \centering
    \caption{Training Time for the First Task on Split-CIFAR-100.}
    \label{tab:training_time_split_cifar100}
    \begin{tabular}{@{}lc@{}}
        \toprule
        \textbf{Method} & \textbf{Training Time (seconds)} \\ \midrule
        ER & 14 \\
        ER + LWF & 15 \\
        MIR & 31 \\
        ER-ACE & 17 \\
        DER & 17 \\
        RAR & 72 \\
        SCR & 131 \\        
        LPR & 213 \\
        \midrule
        OCAR(Ours) & 38 \\
        \bottomrule
    \end{tabular}
\end{table}
OCAR is about 3 times slower than standard ER, which does not perform any additional operations apart from the basic SGD loop. It is also faster than other sophisticated methods such as MIR, SCR, and LPR. We underline that in our tests LPR resulted much slower than what the authors found in their original paper \cite{yoo2024layerwise}. We did not identify the cause of this. It can be something related to our hardware or the implementation details of our code. 

\paragraph{Hyperparameters selection}
As with everything else, the hyper-optimization is performed as in \cite{DBLP:conf/iccvw/Soutif-Cormerais23}. The best configuration is selected as the best performing in the cumulative accuracy metrics on the validation set at the end of the fourth experience. This approach is suboptimal when working on nonstationary data. We notice that in our model, this shorter stream would select a combination of parameters prone to having high plasticity and lower stability. For this reason, instead of selecting a fixed Tikhonov regularizer $\tau$, we instead selected a speed of increase for $\tau$. In this way, we found a much more robust approach. Differently from \cite{DBLP:conf/iccvw/Soutif-Cormerais23}, all our hyperparameters for OCAR have been selected after trying 100 combinations using the tree-structured Parzen estimator algorithm, instead of 200 trials used in the original survey. This can give our method a small disadvantage, but we found it to be sufficient to get good results. The only exception was OCAR-ACE on Split-TinyImageNet where 200 trials were needed due to the particular complexity of using a loss function not derived from a KL divergence on a complex dataset as TinyImageNet. We want to highlight that in the LPR paper \cite{yoo2024layerwise}, it seems the hyperparameters selection is done by training on all the experiences and selecting the best final accuracy. It is possible this can result in an advantage, avoiding the problem of the short and partial stream used in \cite{DBLP:conf/iccvw/Soutif-Cormerais23} and for our method.

\section{Fisher Matrix Computations}
\label{app: Fisher computations}
The Fisher Information Matrix can be computed both as the variance of the score (the gradient of the log-likelihood) or as the negative expected value of the curvature of the log-likelihood. For a classification problem we assume the model $f_w$ is predicting a probability vector probability vector $\mathbf{p}=[p_1, \ldots, p_K]$ where each $p_k$ is the probability that $y_k=1$, such that $\mathbf{p}=f(x;w)$. The distribution assumed is then a categorical distribution. On a single example $x$ the FIM can be computed as:

\begin{align*}
F(w) &=\mathbf{E}_{\mathbf{y}\sim Cat(\mathbf{y}|x;w)}[\nabla_w^2  \, logCat(\mathbf{y}|x;w)] = \mathbf{E}_{\mathbf{y}\sim Cat(\mathbf{y}|x;w)} [\nabla_w^2 \, \sum_k y_k log f_w(x)_k] = \\
&=  \mathbf{E}_{\mathbf{y}\sim Cat(\mathbf{y}|x_n;w)}[
\begin{bmatrix}
           \sum_k \frac{y_{k}}{f_w(x)_k}\frac{\partial f_w(x)_k}{\partial w_1}  \\
           \vdots \\
          \sum_k \frac{y_{k}}{f_w(x)_k}\frac{\partial f_w(x)_k}{\partial w_H} 
\end{bmatrix}
\begin{bmatrix}
           \sum_k \frac{y_{k}}{f_w(x)_k}\frac{\partial f_w(x)_k}{\partial w_1}  \\
           \vdots \\
          \sum_k \frac{y_{k}}{f_w(x)_k}\frac{\partial f_w(x)_k}{\partial w_H} 
\end{bmatrix}^T] = \\
&= \mathbf{E}_{\mathbf{y}\sim Cat(\mathbf{y}|x_n;w)}[
\begin{bmatrix}
           \sum_k \frac{y_{k}^2}{f^2_w(x)_k}(\frac{\partial f_w(x)_k}{\partial w_1})^2 
           &\sum_k \frac{y_{k}^2}{f^2_w(x)_k}\frac{\partial f_w(x)_k}{\partial w_1} \frac{\partial f_w(x)_k}{\partial w_2}
           \ldots \\
          \sum_k \frac{y_{k}^2}{f^2_w(x)_k}\frac{\partial f_w(x)_k}{\partial w_1} \frac{\partial f_w(x)_k}{\partial w_2} \\         
           \vdots \\         
\end{bmatrix}] = \\
&= \begin{bmatrix}
           \sum_k \frac{1}{f_w(x)_k}(\frac{\partial f_w(x)_k}{\partial w_1})^2   
           &\sum_k \frac{1}{f_w(x)_k}\frac{\partial f_w(x)_k}{\partial w_1} \frac{\partial f_w(x)_k}{\partial w_2}
           \ldots \\
          \sum_k \frac{1}{f_w(x)_k}\frac{\partial f_w(x)_k}{\partial w_1} \frac{\partial f_w(x)_k}{\partial w_2} \\
           \vdots \\
\end{bmatrix}
\end{align*}

Note how each element is a sum of the contribution from each class, weighted by the inverse predicted probability for that class. The Fisher computed for a gradient composed of the sum (or mean) of gradients from multiple examples is the sum (or mean) of the individual Fishers. This is due to the required assumption to compute the Fisher: being at the optimum of the log-likelihood optimization, assuming the predicted parameters are the true ones. A consequence of this assumption is that the correlations between gradients of different examples will be zero: $\mathbf{V}[\nabla(\mathbf{y}_i| x_i; w) + \nabla(\mathbf{y}_j| x_j; w)] = \mathbf{V}[\nabla(\mathbf{y}_i| x_i; w) ] + \mathbf{V}[\nabla(\mathbf{y}_j| x_j; w) ]$. Then, our $\lambda$ to give more weight to the buffer data is easily implemented. 
For the same assumption about the optimum, the variance of the score is computed as the expected value of the squared gradients: the squared expected value would be equal to zero.

\paragraph{Last Layer FIM} Focusing on the FIM of the last layer and ignoring all the correlations between the weights for a moment, we can compute the diagonal element of a classifier's weight. Assuming the loss $\mathcal{L}$ is a standard cross entropy loss:
\begin{equation*}
    \frac{\partial \mathcal{L}}{\partial w_{j,i}} = h_j (p_i-y_i)
\end{equation*}
where $p_i$ is the prediction for the class $i$ and $y_i$ an indicator function for the real value. 
The variance of the score for this weight is:
\begin{equation*}
    \mathbf{E}\left[ \left(\frac{\partial \mathcal{L}}{\partial w_{j,i}}\right)^2\right] =  \mathbf{E}[h_j^2 (p_i - y_i)^2] = h_j^2 p_i (1-p_i)
\end{equation*}

\begin{figure}[t]
    \centering
    \includegraphics[width=1\linewidth]{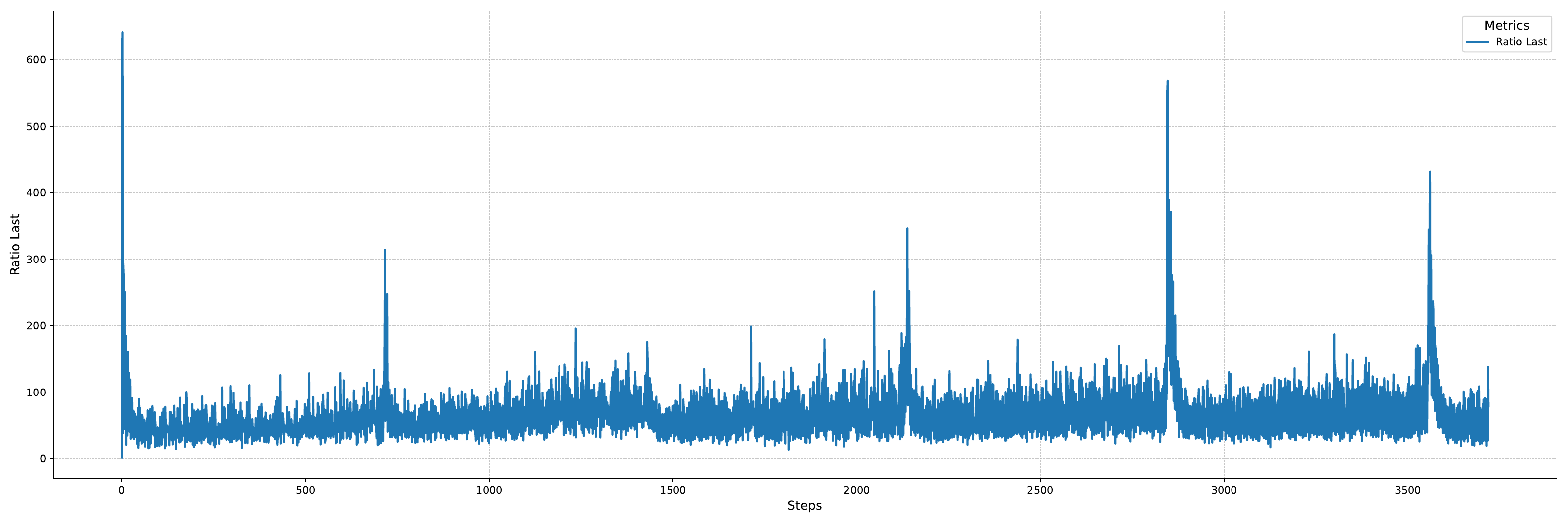}
    \caption{Ratio between the norm of the gradient after being transformed with OCAR and the norm of the original gradient when a small $\tau$ is used}
    \label{fig:ratio_plot}
\end{figure}

If we transform the partial derivative with the inverse variance:
\begin{equation*}
     \mathbf{E}\left[ \left(\frac{\partial \mathcal{L}}{\partial w_{j,i}}\right)^2\right]^{-1} \frac{\partial \mathcal{L}}{\partial w_{j,i}} = \frac{1}{h_j}\frac{p_i - y_i}{p_i (1 - p_i)}
\end{equation*}
In class-incremental settings, at task boundaries, we can assume the probability predicted for the new classes will be pretty low due to the random initialization of the weights. In this case, the gradient of the weights connected to this new unit will be high while the variance of the gradient will be very low, greatly accelerating the gradient, and contributing to possible instabilities due to the assumption of the Fisher of predicting the "true" distribution while, particularly at task boundaries, this is not true. 
Clearly this analysis ignores all the complex cross-correlations between the weights. To visualize the complete effect, we plot in Figure \ref{fig:ratio_plot} the ratio between the norm of the gradient after being transformed with OCAR and the norm of the original cross-entropy gradient. We show only a subset of the training on Split-CIFAR100, using a small $\tau$ (lr/100). All task boundaries are clearly visible, showing the great acceleration the FIM provides when new classes arrive. This underlines the importance of the Tikhonov regularization ($\tau$) in class-incremental settings, where the Fisher can exacerbate some instabilities when the predictions of the model are completely off. 

\section{Qualitative Experiment}
\label{app: qualitative}
The code of the qualitative experiment is included in our repository. We used a basic linear model with 10 inputs and a single output trained to solve a simple linear regression problem with an MSE loss. 10 tasks with 1000 samples per task are randomly generated. Each task has a different multivariate normal from which input data are sampled and different real weights that should be estimated by the mode. The data are accessed as a stream, with a single pass per batch. The  Standard Vitter algorithm is used to keep a buffer of old data. Each method uses a batch composed of 10 new data and 10 data sampled from the buffer. 

Natural Gradient is applied directly by estimating the full Fisher information of the model with an EMA of the FIM computed on the single batches. This FIM after the EMA is used to precondition the gradient. 

OCAR uses the exact same approach but uses a $\lambda$ to give more weight to old data and uses a scheduling of both $\lambda$ and $\tau$ (the Tikhonov regularize) to increase them both in time.
EWC is more tricky to implement in its basic form for online problems. Our approach is very similar to what is done in \cite{mai2022online}, an online extension of the EWC++ strategy \cite{chaudhry2018riemannian}. The difference is that we use also replay data for computing the gradient. Namely, the loss is computed on both old and new data, but the Fisher (being a penalization) is computed only on old data (an EMA is kept also in this case for estimation). Given that no task boundaries are accessible, the penalization of the movement of the weights is done with respect to the weights at the previous step in time. In this way, a more regularized descent should be followed, penalizing the displacement from step to step using the Fisher Information of old data. 
The loss is a basic MSE but for analysis purposes, the figure \ref{app: qualitative} shows the cumulative loss (simply the loss experienced in each batch accumulated in time) and the cumulative loss on all previous data (at each step the loss on all previous data is computed after the optimization step and accumulated in time). The first measure shows a general capability of fast adaptation while the second an ability to find an optimum stable for the entire previous stream.

A hyperparameter selection is performed to select the best setting on the sum of the final value of both these stability and plasticity measures. Then, on the same data, using the best hyperparameters, the results are averaged across 10 random seeds to test fro training stability with different optimization paths.

\section{Online CLEAR}
\label{app:clear}
Following the very recent work of \cite{yoo2024layerwise}, we tested our method also on the Online CLEAR benchmark \cite{lin2021clear}, a domain incremental learning scenario. This scenario is fundamentally different from the class-incremental, with the same classes that undergo some sort of evolution. The CL aspect is then less impactful, with much more forward/backward transfer and less catastrophic forgetting due to the similarity of the tasks in different domains. 
We followed the same settings as in LPR paper \cite{yoo2024layerwise}, that are very similar to the standard approach used for our main experiments, but with some differences: the use of the full ResNet18 instead of the Slim version and 10 gradient steps per batch. Unfortunately, we encountered some bugs when their exact code was used in ours, requiring some slight modifications. Not being able to ensure the exact same conditions, we rerun all the experiments. We decided to compare the baseline of Experience Replay, the very recent LPR approach, and our method. ER-ACE is tailored for task-incremental settings, so we avoid its use. In this scenario, we increase $\lambda$ not with the number of classes in the buffer (as in class incremental), but we increase it in time as new batches arrive. LPR was tested using 100 samples from the buffer to estimate its preconditioner.

\begin{figure*}[t!]
    \centering
    \begin{subfigure}[t]{0.18\textwidth}
        \centering
        \includegraphics[width=\textwidth]{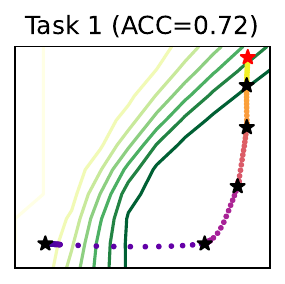}
        \caption{OCAR - T1.}
    \end{subfigure}
    \begin{subfigure}[t]{0.18\textwidth}
        \centering
        \includegraphics[width=\textwidth]{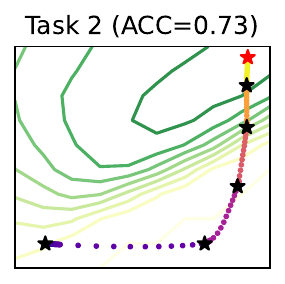}
        \caption{OCAR - T2.}
    \end{subfigure}
    \begin{subfigure}[t]{0.18\textwidth}
        \centering
        \includegraphics[width=\textwidth]{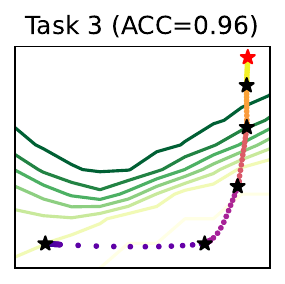}
        \caption{OCAR - T3.}
    \end{subfigure}
    \begin{subfigure}[t]{0.18\textwidth}
        \centering
        \includegraphics[width=\textwidth]{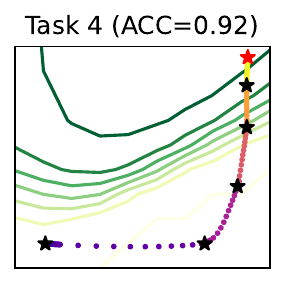}
        \caption{OCAR - T4.}
    \end{subfigure}
    \begin{subfigure}[t]{0.18\textwidth}
        \centering
        \includegraphics[width=\textwidth]{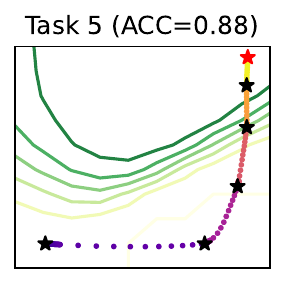}
        \caption{OCAR - T5.}
    \end{subfigure}
    \begin{subfigure}[t]{0.18\textwidth}
        \centering
        \includegraphics[width=\textwidth]{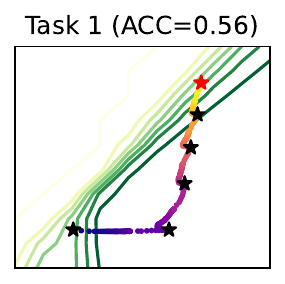}
        \caption{ER - T1.}
    \end{subfigure}
    \begin{subfigure}[t]{0.18\textwidth}
        \centering
        \includegraphics[width=\textwidth]{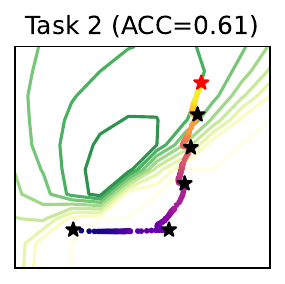}
        \caption{ER - T2.}
    \end{subfigure}
    \begin{subfigure}[t]{0.18\textwidth}
        \centering
        \includegraphics[width=\textwidth]{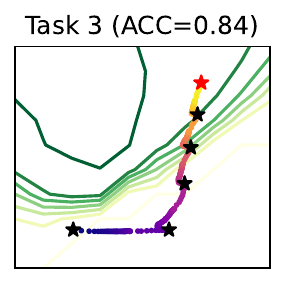}
        \caption{ER - T3.}
    \end{subfigure}
    \begin{subfigure}[t]{0.18\textwidth}
        \centering
        \includegraphics[width=\textwidth]{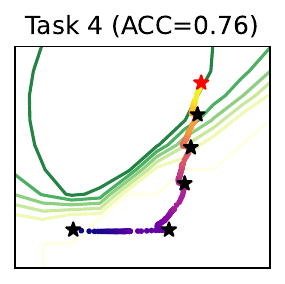}
        \caption{ER - T4.}
    \end{subfigure}
    \begin{subfigure}[t]{0.18\textwidth}
        \centering
        \includegraphics[width=\textwidth]{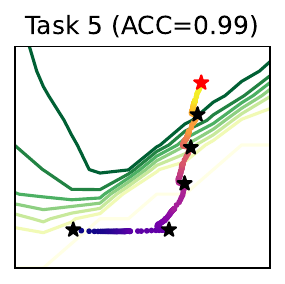}
        \caption{ER - T5.}
    \end{subfigure}
   \begin{subfigure}[t]{0.18\textwidth}
        \centering
        \includegraphics[width=\textwidth]{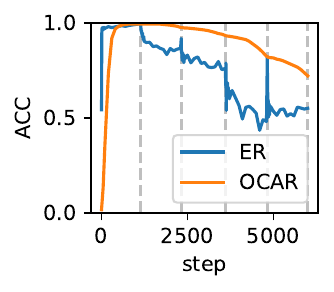}
        \caption{T1.}
    \end{subfigure}
   \begin{subfigure}[t]{0.18\textwidth}
        \centering
        \includegraphics[width=\textwidth]{figs/manifold/learning_curve_T0.pdf}
        \caption{T2.}
    \end{subfigure}
   \begin{subfigure}[t]{0.18\textwidth}
        \centering
        \includegraphics[width=\textwidth]{figs/manifold/learning_curve_T0.pdf}
        \caption{T3.}
    \end{subfigure}
   \begin{subfigure}[t]{0.18\textwidth}
        \centering
        \includegraphics[width=\textwidth]{figs/manifold/learning_curve_T0.pdf}
        \caption{T4.}
    \end{subfigure}
   \begin{subfigure}[t]{0.18\textwidth}
        \centering
        \includegraphics[width=\textwidth]{figs/manifold/learning_curve_T0.pdf}
        \caption{T5.}
    \end{subfigure}
   
    \caption{2D projections of the training trajectories for ER and OCAR on Split MNIST (5 Tasks). The black stars highlight the task boundaries, the red star the final model. We also show learning curves on each task separately. }
    \label{fig:app_traj}
\end{figure*}

\section{Learning Trajectory on Split MNIST}
\label{app: trajectory}

The model is a small feedforward network with two hidden layers of $100$ units and ReLU activations. The model is trained online with 3 passes for each mini-batch with a small replay buffer of $100$ elements, corresponding to $10$ samples per class by the end of training.

To plot the 2D learning projections in Figure \ref{fig:app_traj}, we follow a procedure similar to \cite{DBLP:conf/iclr/MirzadehFGP021}.

We consider the 2D plane that intersects the model's initialization $w_0^*$, the model after the first task $w_1^*$, and the final model $w_5^*$. Many other possibilities were considered before this choice (e.g. using different tasks or random directions), but they all resulted in qualitatively similar plots. The coordinates system is obtained by orthonormalizing the directions $u = w_1^* - w_0^*$ and $v = w_5^* - w_0^*$, obtaining $\bar{u} = \frac{u}{\left\lVert u \right\rVert}$ and $\bar{v} = \frac{v - \cos(u, v) u}{\left\lVert v - \cos(u, v) u \right\rVert}$. Given a model $w$, its coordinates in the 2D space are the unique $\langle x, y \rangle$ such that $w = x \bar{u} + y \bar{v} + w_0^*$. Notice that each method has different values for $w_1^*$ and $w_5^*$, and therefore different 2D planes and coordinate systems are chosen for each method. Since the coordinates are not meaningful and cannot be compared across plots, we do not show them in Figure \ref{fig:app_traj}.

\section{Additional Figures}
\label{app: figures}
\begin{figure}[h!]
    \centering
    \includegraphics[width=0.7\linewidth]{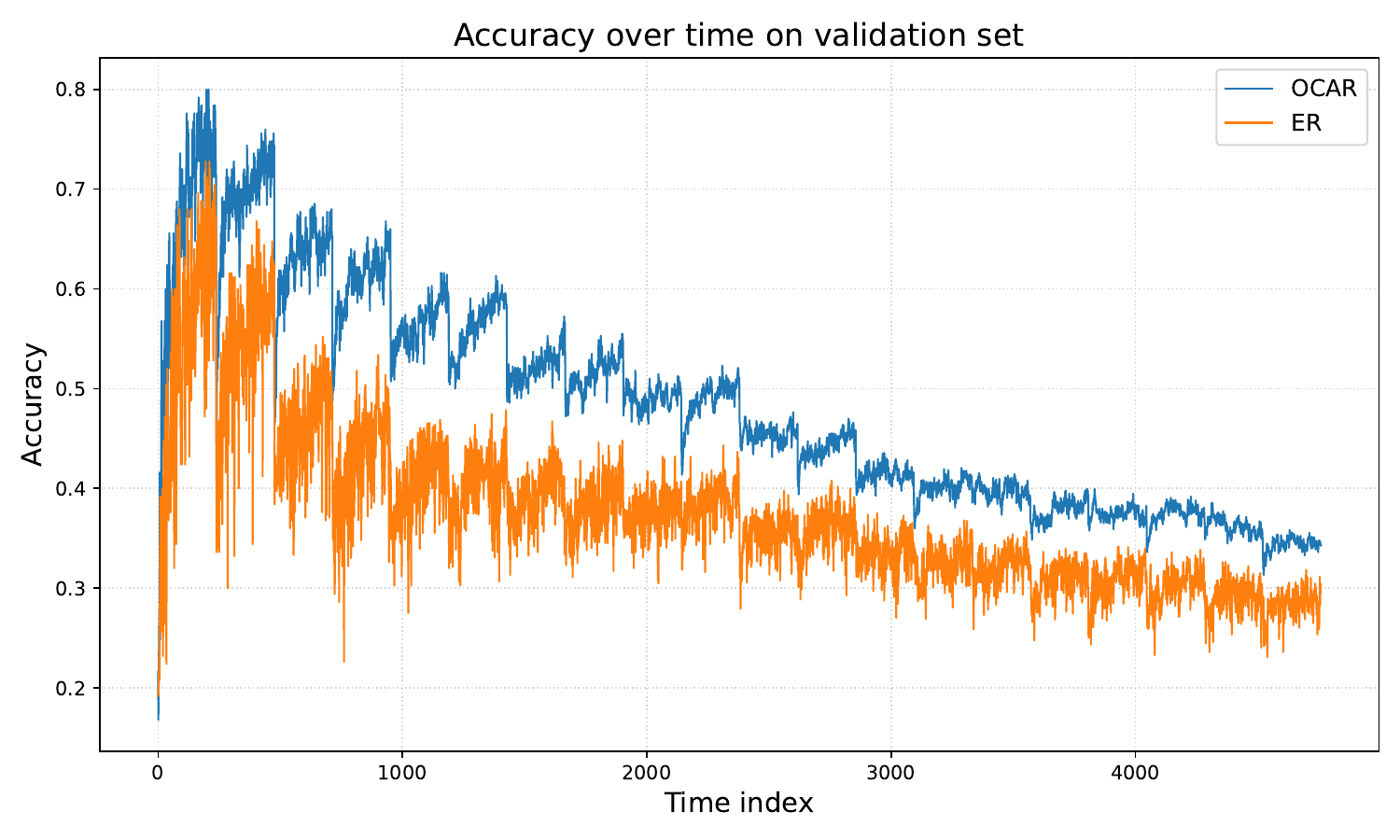}
    \caption{Accuracy over time on the validation set for Split-CIFAR100 experiment.}
    \label{fig:enter-label}
\end{figure}
\begin{figure}[h!]
    \centering
    \includegraphics[width=0.7\linewidth]{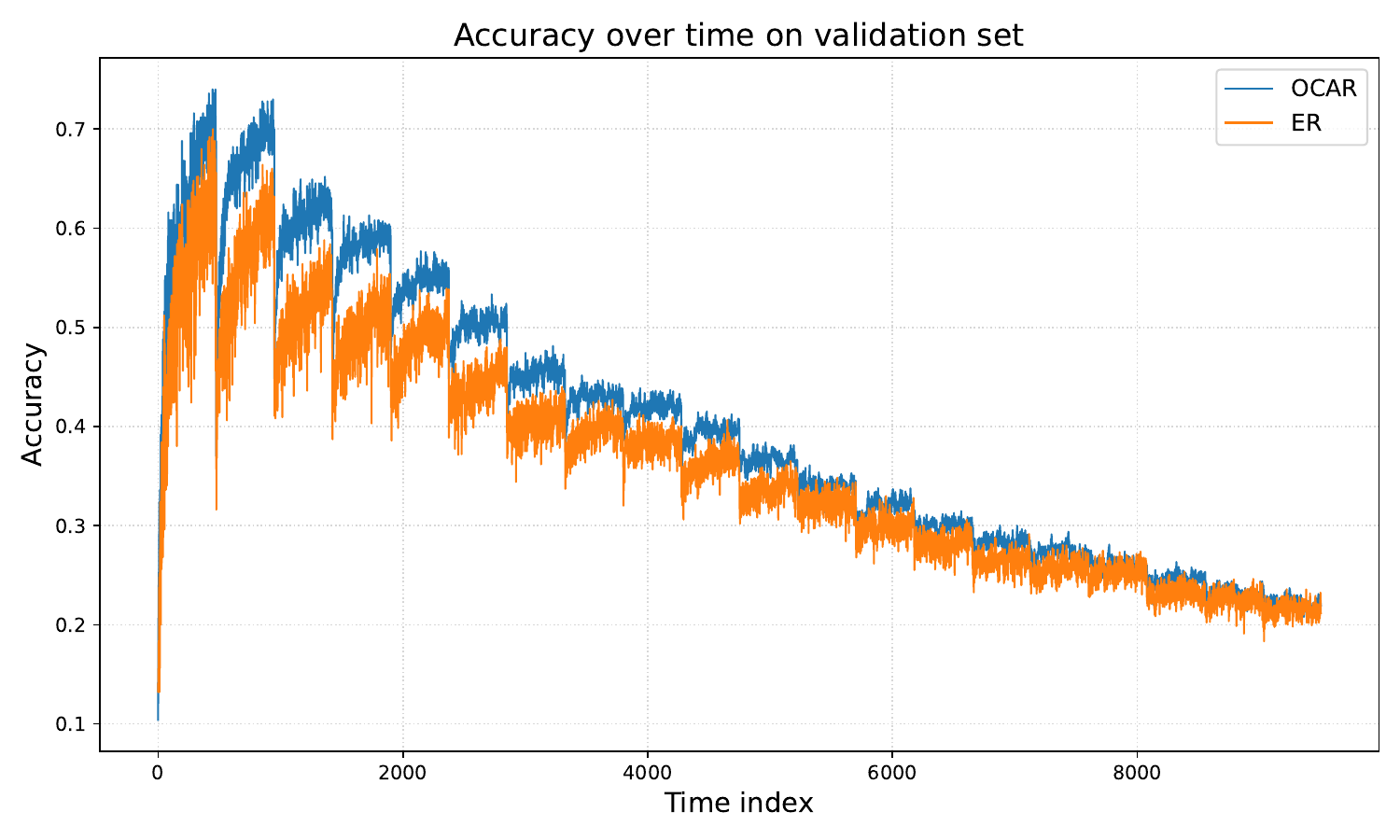}
    \caption{Accuracy over time on the validation set for Split-TinyImagenet experiment.}
    \label{fig:enter-label}
\end{figure}

\begin{figure}
    \centering
    \includegraphics[width=0.7\linewidth]{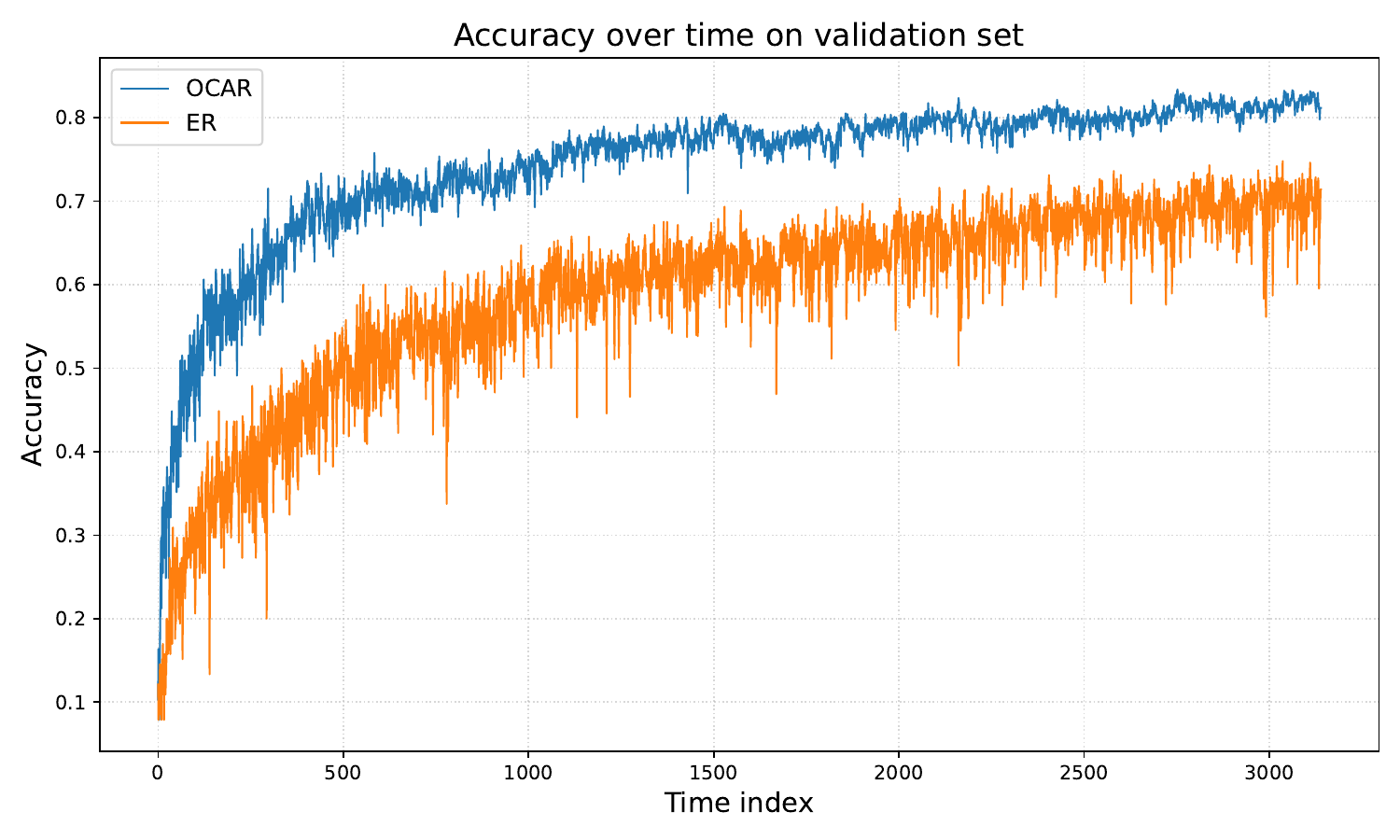}
    \caption{Accuracy over time on the validation set for Online CLEAR experiment.}
    \label{fig:clear accuracy}
\end{figure}

\begin{figure}[!h]
    \centering
    \includegraphics[width=0.5\textwidth]{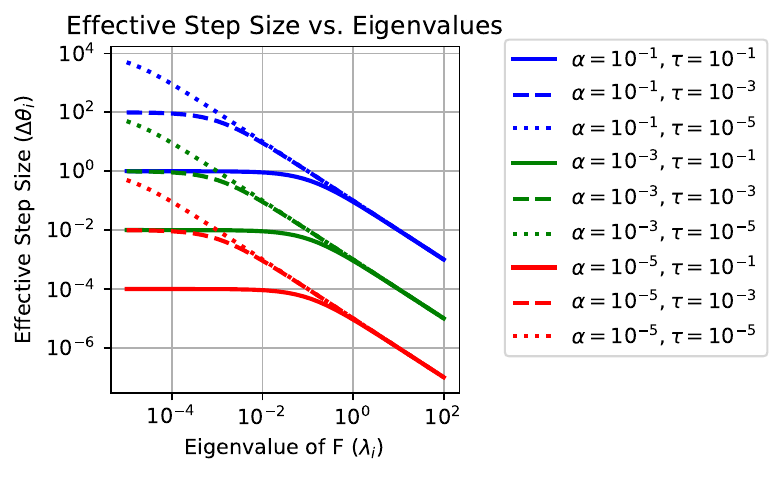}
    \caption{Effective step size against the FIM eigenvalues. The step size in the directions with small eigenvalues is regularized via $\tau$, while large eigenvalues are unaffected by it. The learning rate $\alpha$ affects all the directions equally.}
    \label{fig:lr_ratio_curve}
\end{figure}

\begin{figure}[t]
    \centering
    \begin{subfigure}[t]{0.3\textwidth}
        \centering
        \includegraphics[width=\textwidth]{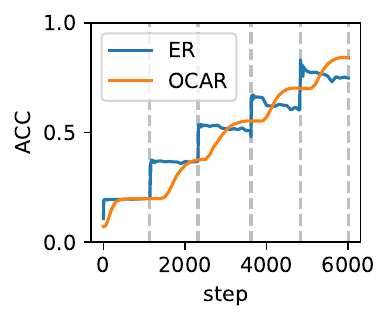}
        \caption{Average of all tasks.}
        \label{fig:curve_joint}
    \end{subfigure}   
    \begin{subfigure}[t]{0.3\textwidth}
        \centering
        \includegraphics[width=\textwidth]{figs/manifold/learning_curve_T0.pdf}
        \caption{First Task.}
        \label{fig:curve_t0}
    \end{subfigure}
    
    \caption{Learning curves on the first task (\ref{fig:curve_t0}) and the average accuracy for all tasks (\ref{fig:curve_joint}). }\label{fig:lr_curves}
\end{figure}

\begin{figure*}[t]
    \centering
    \begin{subfigure}[b]{0.32\textwidth}
        \centering
        \includegraphics[width=\textwidth]{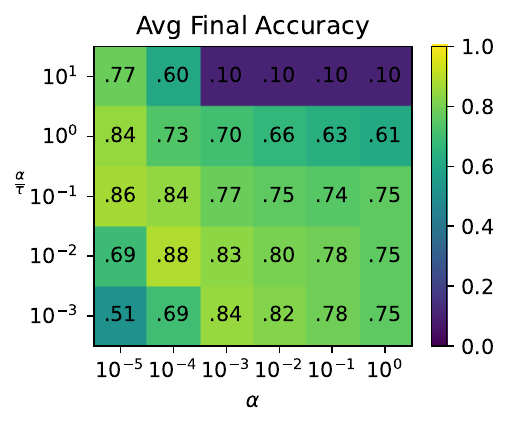}
    \end{subfigure}
    \begin{subfigure}[b]{0.32\textwidth}
        \centering
        \includegraphics[width=\textwidth]{figs/grid_forget.pdf}
    \end{subfigure}
    \begin{subfigure}[b]{0.32\textwidth}
        \centering
        \includegraphics[width=\textwidth]{figs/grid_plast.pdf}
    \end{subfigure}

    \caption{Grid search over $\alpha$ and $\frac{\alpha}{\tau}$: (left) average accuracy, (middle) forgetting on the first task, (right) plasticity measured as the accuracy on the final task. Metrics are computed on the test stream at the end of training.}\label{fig:lr_ratio_grid_complete}
\end{figure*}


\end{document}